\documentclass{article}

\usepackage{microtype}
\usepackage{graphicx}
\usepackage{subcaption}
\usepackage{booktabs} 

\usepackage{hyperref}

\usepackage[preprint]{icml2026}

\usepackage{bm}       
\usepackage{booktabs} 
\usepackage{graphicx}
\usepackage{amsmath}
\usepackage{amssymb}
\usepackage{algorithm}
\usepackage{algorithmic}
\usepackage{mathtools}
\usepackage{amsthm}
\usepackage{multirow}
\usepackage{array}
\usepackage{color}
\usepackage{xcolor} 
\usepackage{graphicx} 
\usepackage{caption}
\usepackage[capitalize,noabbrev]{cleveref}

\theoremstyle{plain}

\theoremstyle{definition}

\theoremstyle{remark}

\usepackage[table]{xcolor}
\usepackage{wrapfig}
\usepackage{pifont}
\newcommand{\ie}{\textit{i}.\textit{e}.}
\newcommand{\eg}{\textit{e}.\textit{g}.}

\definecolor{lightgray}{gray}{0.9}
\definecolor{darkgreen}{RGB}{0, 150, 0}
\definecolor{myblue}{RGB}{15,142,208}
\newcommand{\gain}[1]{{\textcolor{darkgreen}{\scriptsize#1}}}

\newcolumntype{L}{>{\columncolor{lightgray}}l}

\definecolor{mygray}{gray}{.9}
\usepackage[textsize=tiny]{todonotes}
\newcommand{\myhyperlink}[3][black]{\hyperlink{#2}{\color{#1}{#3}}}

\usepackage{xspace}
\makeatletter
\DeclareRobustCommand\onedot{\futurelet\@let@token\@onedot}
\def\@onedot{\ifx\@let@token.\else.\null\fi\xspace}
\def\eg{\emph{e.g}\onedot} 
\def\ie{\emph{i.e}\onedot} 
 
 \def\vs{\emph{vs}\onedot}

\newcommand{\thickhline}{%
    \noalign {\ifnum 0=`}\fi \hrule height 1pt
    \futurelet \reserved@a \@xhline
}

\makeatletter
\def\hlinewd#1{%
  \noalign{\ifnum0=`}\fi\hrule \@height #1 \futurelet \reserved@a\@xhline}
\makeatother
\icmltitlerunning{Path-Decoupled Hyperbolic Flow Matching for Few-Shot Adaptation}

\begin{document}

\twocolumn[
  \icmltitle{Path-Decoupled Hyperbolic Flow Matching for Few-Shot Adaptation}

  \icmlsetsymbol{equal}{*}

  \begin{icmlauthorlist}
    \icmlauthor{Lin Li}{yyy}
    \icmlauthor{Ziqi Jiang}{yyy}
    \icmlauthor{Gefan Ye}{comp}
    \icmlauthor{Zhenqi He}{yyy}
    \icmlauthor{Jiahui Li}{comp}
    \icmlauthor{Jun Xiao}{comp}
    \icmlauthor{Kwang-Ting Cheng}{yyy}
    \icmlauthor{Long Chen}{yyy}
  \end{icmlauthorlist}

  \icmlaffiliation{yyy}{The Hong Kong University of Science and Technology (HKUST)}
  \icmlaffiliation{comp}{Zhejiang University}

  \icmlcorrespondingauthor{Long Chen}{longchen@ust.hk}

  \icmlkeywords{Machine Learning, ICML}

  \vskip 0.3in
]



\printAffiliationsAndNotice{}  

\begin{abstract}
Recent advances in cross-modal few-shot adaptation treat visual-semantic alignment as a continuous feature transport problem via Flow Matching (FM). However, we argue that Euclidean-based FM overlooks fundamental limitations of flat geometry, where polynomial volume growth fails to accommodate diverse feature distributions, leading to severe \textbf{path entanglement}. To this end, we propose path-decoupled Hyperbolic Flow Matching (HFM), leveraging the Lorentz manifold's exponential expansion for trajectory decoupling. HFM structures the transport via two key designs: 1) \textit{Centripetal hyperbolic alignment}: It constructs a centripetal hierarchy by anchoring textual roots, which pushes visual leaves to the boundary to initialize orderly flows. 2) \textit{Path-decoupled objective}: It acts as a ``semantic guardrail'' rigidly confining trajectories within isolated class-specific geodesic corridors via step-wise supervision. Furthermore, we devise an adaptive \textit{diameter-based stopping} to prevent over-transportation into the crowded origin based on the intrinsic semantic scale. Extensive ablations on 11 benchmarks have shown that HFM establishes a new state-of-the-art, consistently outperforming its Euclidean counterparts. Our codes and models will be released.

\end{abstract}

\section{Introduction}
The remarkable zero-shot generalization of pretrained vision-language models (VLMs), such as CLIP~\cite{radford2021learning}, builds upon their ability to embed images and natural language into a shared semantic embedding space~\cite{jia2021scaling,li2022grounded,zhong2022regionclip}. Although such joint representations facilitate zero-shot classification via cross-modal similarity, a large performance gap remains when encountering specialized downstream tasks~\cite{zhou2022learning,zhang2022tip}. To bridge this gap, few-shot adaptation~\cite{zhou2022learning,zhou2022conditional,zhang2022tip,gao2024clip} aims to refine the pre-trained encoders, leveraging minimal supervisory signals to realign visual features with their corresponding textual prototypes.

\begin{figure}
    \centering
    \includegraphics[width=1.0\linewidth]{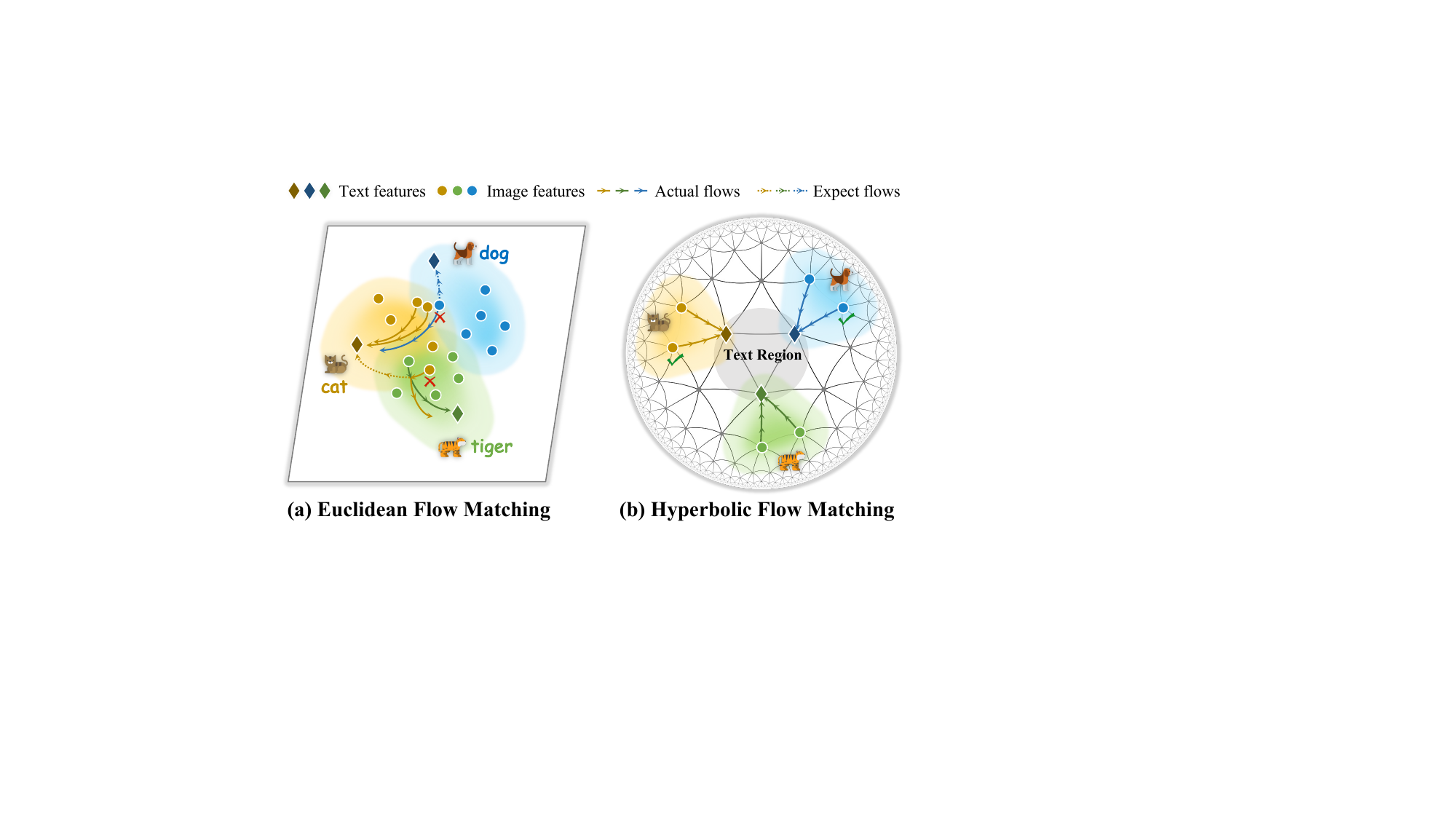}
    \put(-173, 81.5){\tiny{\myhyperlink{Q2}{\ding{183}}}}
    \put(-176.2, 56.8){\tiny{\myhyperlink{Q1}{\ding{182}}}}
    \caption{Illustration of Path Entanglement. (a) Euclidean Flow Matching suffers from severe trajectory collisions (\eg, ``cat'' intersecting ``tiger'' and ``dog'' merging with ``cat'') due to the limited polynomial capacity of flat geometry. (b) Hyperbolic Flow Matching uses exponential volume expansion to achieve path decoupling.}
    \label{fig:intro}
    \vspace{-1.5em}
\end{figure}

To achieve such adaptation, various Parameter-Efficient Fine-Tuning (PEFT) techniques, \eg, prompt tuning~\cite{zhou2022learning,zhou2022conditional} and feature-level adapters~\cite{gao2024clip,hu2022lora,zanella2024low}, have been widely adopted to refine latent representations. However, these methods predominantly rely on a ``one-step'' adjustment, where the alignment is restricted to a single forward pass of the adapter module. Such a direct transformation often struggles to resolve complex semantic entanglements in challenging datasets, due to the lack of the iterative rectification inherent in continuous processes~\cite{jiang2025exploring}. To address this, recent efforts like FMA~\cite{jiang2025exploring} treat alignment as a continuous feature transport problem via Flow Matching (FM)~\cite{lipman2022flow, liu2022flow} strategy. Specifically, they construct conditional flow paths that guide visual features from the image manifold to the corresponding class-specific text regions, thereby eliminating the cross-modal gap for classification. By learning a velocity field that defines a time-continuous evolution, the aforementioned methods enable multi-step rectification. This iterative refinement allows the model to progressively correct alignment errors, enabling superior expressive capacity over one-step counterparts.

Despite the theoretical flexibility of multi-step flows, we argue that existing Euclidean FM models suffer from severe inherent \textbf{path entanglement}. It refers to the transport paths that inadvertently intersect, overlap, or merge within the latent space. Fundamentally, this entanglement arises because the polynomial volume growth of flat geometry fails to accommodate diverse feature distributions, leading to two inevitable structural interferences: \myhyperlink{Q1}{\ding{182}} \textit{\textbf{Disordered Cross-Modality Flows}}. Source image features and their corresponding target semantic prototype are embedded in \textit{irregularly dispersed} positions. Bridging this unstructured gap necessitates long-range transport, thereby increasing the risk of trajectory collisions (\eg, the yellow flow of the ``cat'' colliding with and following the green path of the ``tiger'' in Figure~\ref{fig:intro}a). This chaotic confluence erodes feature discriminability, thereby compromising the classification performance. \myhyperlink{Q2}{\ding{183}} \textit{\textbf{Crowded Inter-Class Flows}}. Flow sources from different categories may unintentionally overlap, leading to more ambiguous transport trajectories. Without sufficient separation, these paths are easily diverted by \textit{high-density neighboring clusters} (\eg, the blue flow of ``dog'' drifting into the dense yellow samples in Figure~\ref{fig:intro}a).

To this end, in this paper, we propose a new path-decoupled \textbf{H}yperbolic \textbf{F}low \textbf{M}atching (\textbf{HFM}) for few-shot adaptation. By reformulating the transport dynamics within the Lorentz manifold, we exploit the exponential volume growth of hyperbolic geometry to spatially decouple the transport trajectories to alleviate path entanglement. To address \myhyperlink{Q1}{\ding{182}}, we employ a mechanism of \textbf{centripetal hyperbolic alignment}. It restructures the latent geometry into a concentric hierarchy by explicitly initializing textual prototypes with small hyperbolic radii (proximal to the origin, visualized as the central ``Text Region'' in Figure~\ref{fig:intro}b) and visual features near the manifold boundary. To enforce this \textit{centripetal order}, we leverage the \textit{cross-modality entailment} objective~\cite{desai2023hyperbolic} to optimize the embedding space, formally establishing text as the semantic root and images as entailment leaves. By constraining centripetal order flows to focusing radial geodesics, we can reduce trajectory collisions, thereby preserving semantic discriminability and enhancing classification performance. To tackle \myhyperlink{Q2}{\ding{183}}, we introduce a \textbf{path-decoupled objective}. Leveraging the exponential volume expansion near the manifold boundary, this objective maximizes the geodesic margin between distinct semantic categories. This optimization explicitly constructs isolated geodesic transport corridors (illustrated as the separated yellow, green, and blue areas in Figure~\ref{fig:intro}b), ensuring that each class-specific flow evolves within a unique, non-overlapping volume. Consequently, this geometric separation effectively decouples inter-class trajectories, preventing the ambiguous overlaps inherent to crowded Euclidean spaces. 

Furthermore, in the inference stage, to enhance inference efficiency and accuracy, we devise an adaptive \textbf{diameter-based stopping} strategy. By dynamically terminating the flow upon the convergence of geometric compactness, we prevent over-transportation into the densely populated text region near the origin. This avoids the critical risk of visual features drifting into adjacent, incorrect clusters due to spatial crowding, ensuring both precise classification and minimal computational overhead.

Extensive experiments on 11 few-shot benchmarks demonstrate that our HFM consistently surpasses state-of-the-art Euclidean-based FM counterparts and PEFT adapters as a plug-and-play module. These promising results evidence the significant potential of non-Euclidean manifolds for advancing the frontier of few-shot cross-modal understanding.
\section{Related Works}

\textbf{Few-shot Adaptions in VLMs.} Few-shot adaptation is a task that requires models to learn from only a few annotated examples~\cite{chen2020closer, hou2019cross, wang2020generalizing}. Current approaches in VLMs usually follow the paradigm that employs pre-trained VLMs and then refines them through task-adaptive optimization. Parameter-Efficient Fine-Tuning (PEFT)~\cite{gu2022ppt, hu2023scaled, lee2023read, zhang2022tip, zanella2024low, de2025take} is one of popular optimization methods, which focuses on updating a small subset of parameters to achieve performance comparable to full fine-tuning. In the meanwhile, some studies explore the potential of in-context learning~\cite{alayrac2022flamingo, zhang2023multimodal, hu2022context, cai2023context, huang2024select} in few-shot adaptation. With no change on model parameters, they add support examples into the inference context to make pre-trained models adapt to specific tasks. Despite computational efficiency and effectiveness, their one-step adjustment fails to disentangle complex feature distributions, leaving substantial room for improvement.


\noindent\textbf{Hyperbolic Representation Learning.} Hyperbolic geometry, such as Poincar{\'e}-ball Model~\cite{nickel2017poincare} and Lorentz Model~\cite{nickel2018learning}, demonstrates extraordinary capability in modeling hierarchical relationships due to its exponential volume growth relative to radius. With custom-designed neural network modules~\cite{chen2022fully, liu2019hyperbolic, li2025text}, hyperbolic representation learning has achieved profound breakthroughs in a wide range of tasks in different modalities, including text~\cite{dhingra2018embedding, zhu2020hypertext, le2019inferring}, image~\cite{atigh2022hyperbolic, khrulkov2020hyperbolic, li2023euclidean}, audio~\cite{petermann2023hyperbolic, nakashima2022hyperbolic} and video~\cite{li2025hlformer, li2025enhancing}. Meanwhile, combining with contrastive learning methods enables hyperbolic representation learning to successfully adapt to cross-modal tasks~\cite{wen2025hover, desai2023hyperbolic, liu2020hyperbolic, hong2023hyperbolic}. Given the inherent cross-modality hierarchy of the data in few-shot adaptation task, we try to harness the power of hyperbolic geometry for flow matching.

\textbf{Flow Matching (FM).} Diffusion Model~\cite{ho2020denoising, song2020score, rombach2022high} is a powerful generative tool, but limited by slow inference due to curved sampling paths. Flow Matching~\cite{lipman2022flow, liu2022flow} addresses this limitation by learning vector fields that generate straight-line trajectories between distributions. Specifically, the utilization of optimal transport simplifies the geometric structure of the vector field, thus facilitating faster and more accurate sampling. From the perspective of stochastic interpolants, FM and diffusion models act as special cases within a unifying framework~\cite{albergo2023stochastic}. In spite of remarkable successes in domains like image generation~\cite{ren2024flowar, esser2024scaling, luo2025curveflow}, FM has recently been introduced to cross-modal adaptation. Notable works like FMA~\cite{jiang2025exploring} treat visual-semantic alignment as a continuous feature transport problem. However, constrained by flat geometry, these Euclidean-based FM often suffer from path entanglement, impairing the classification performance.

\section{Approach}
\label{sec:method}
\begin{figure*}
    \centering
    \includegraphics[width=1\linewidth]{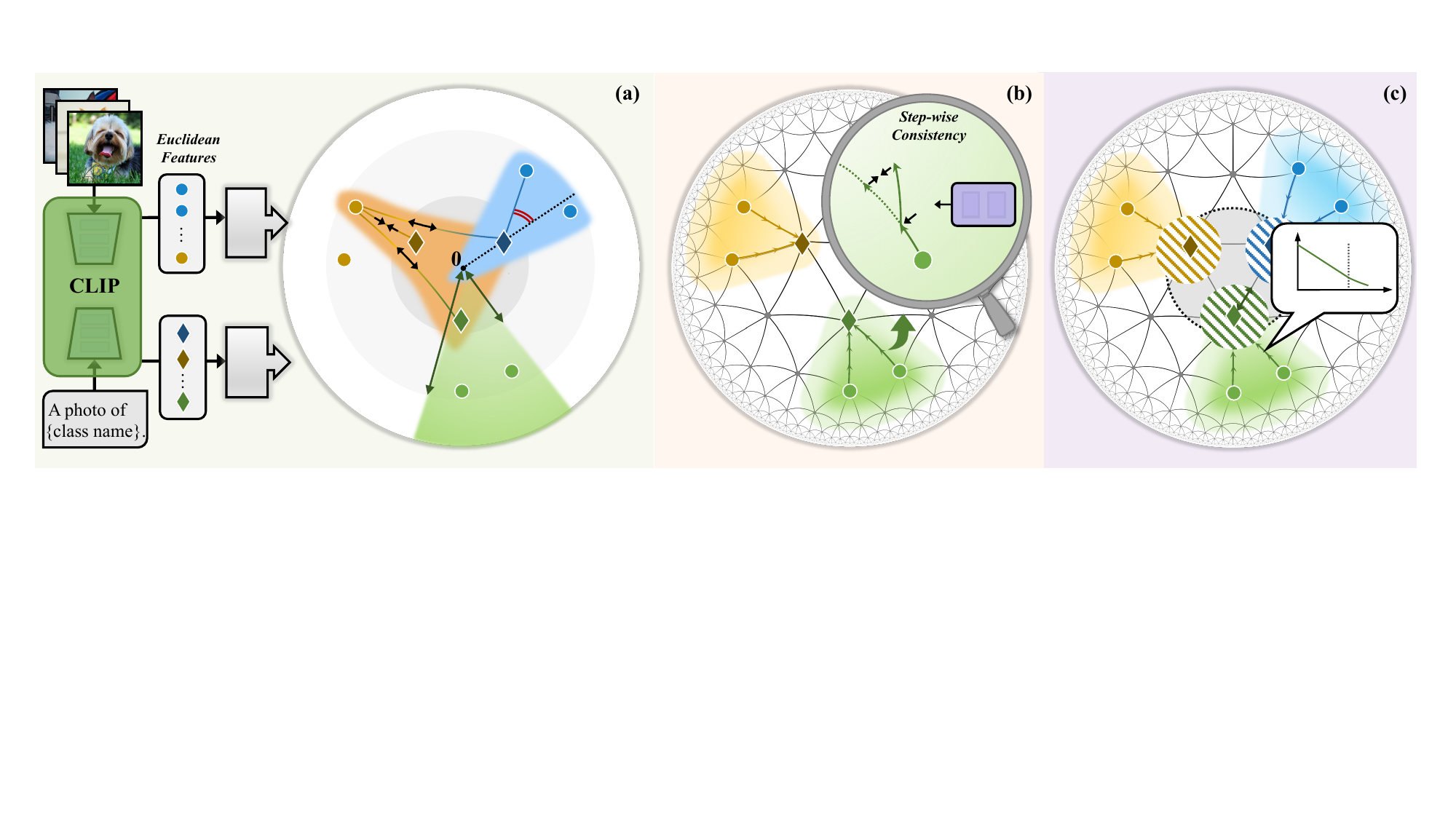}
    \put(-419, 83.5){\scalebox{0.6}{$ \text{exp}_{\bm{0}}$}}
    \put(-419, 34){\scalebox{0.6}{$ \text{exp}_{\bm{0}}$}}
    \put(-308, 99){\scalebox{0.7}{$\mathcal{L}_{\text{entail}}$}}
    \put(-380, 95){\scalebox{0.7}{$\mathcal{L}_{\text{con}}$}}
    \put(-358, 45){\scalebox{0.7}{$\alpha_{\text{img}}$}}
    \put(-328, 60){\scalebox{0.7}{$\alpha_{\text{txt}}$}}
    \put(-314, 89){\scalebox{0.7}{$\omega$}}
    \put(-177, 63){\scalebox{0.7}{$\bm{x}_0$}}
    \put(-191, 80){\scalebox{0.7}{$\bm{x}_t$}}
    \put(-186, 107){\scalebox{0.7}{$\hat{\bm{x}}_{t+\delta}$}}
    \put(-205, 91){\scalebox{0.7}{$\bm{x}_{t+\delta}$}}
    \put(-157, 88){\scalebox{0.7}{$\mathcal{F}_{\theta}$}}
    \put(-177.5, 88.5){\scalebox{0.7}{$v_{t}$}}
    \put(-23, 55){\scalebox{0.6}{time~$t$}}
    \put(-48, 75){\scalebox{0.6}{$d_{\scriptscriptstyle\mathbb{L}}$}}
    \put(-20, 70){\scalebox{0.6}{$\leftarrow$$t^{*}$}}
    \put(-83, 57){\scalebox{0.5}{$\phi(N)\cdot{d}_{\text{txt}}$}}
    \caption{The overview of HFM.
(a)~\textbf{Constructing Centripetal Hyperbolic Space}: Establish a centripetal cross-modal hierarchy, optimizing textual roots near the origin and visual leaves toward the boundary.
(b)~\textbf{Learning Path-Decoupled Flows}: Tangent velocity fields $\mathcal{F}_\theta$ are optimized to guide features along isolated geodesic corridors. The step-wise consistency enforces trajectory decoupling in hyperbolic space.
(c)~\textbf{Inference with Diameter-based Stopping}: The flow terminates at $t^*$ once the distance to text prototypes drops below a dynamic threshold scaled by semantic diameter $d_{\text{txt}}$ to alleviate over-transportation into the crowded
origin.}
\label{fig:framework}
\vspace{-1.0em}
\end{figure*}
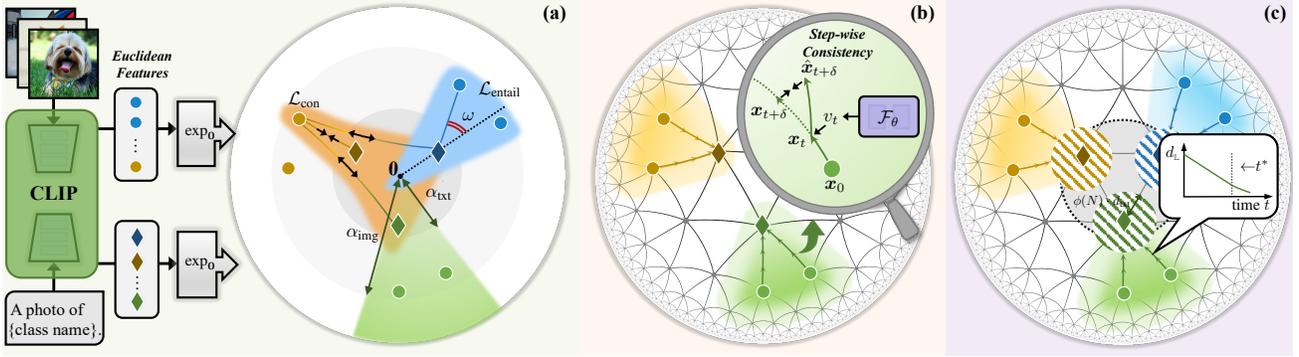
In this section, we present HFM, a path-decoupled framework designed to alleviate the path entanglement in few-shot adaptation. As illustrated in Figure~\ref{fig:framework}, our HFM comprises three phases: 
1)~\textbf{Constructing Centripetal Hyperbolic Space}~(\S\ref{sec:centripetal}): Establishing a centripetal hierarchy by anchoring textual roots at the origin and visual leaves at the boundary.
2)~\textbf{Learning Path-Decoupled Flows}~(\S\ref{sec:flow_matching}): Optimizing a tangent velocity field to confine transport within isolated geodesic corridors via step-wise and contrastive supervision. 3)~\textbf{Inference with Diameter-based Stopping}~(\S\ref{sec:inference}): Employing a density-aware termination criterion to prevent over-transportation in crowded text region.

\subsection{Preliminaries: Lorentz Model}
We first briefly introduce the key concepts of hyperbolic geometry on which our HFM relies. Following~\citep{desai2023hyperbolic,pal2025compositional}, we adopt the Lorentz model of hyperbolic space $\mathbb{L}^{n,\kappa}$ with \textit{learnable} constant negative curvature $-\kappa$ ($\kappa > 0$). In this model, each point $\bm{x} \in \mathbb{L}^{n, \kappa}$ consists of a time-like scalar $x_0$ and a space-like vector $\tilde{\bm{x}}$, constrained by the geometric property $\langle \bm{x}, \bm{x} \rangle_{\scriptscriptstyle\mathbb{L}} = -1/\kappa$. The Lorentzian inner product is formulated as
$\langle \bm{x}, \bm{y} \rangle_{\scriptscriptstyle\mathbb{L}} = - x_0 y_0 + \langle \tilde{\bm{x}}, \tilde{\bm{y}} \rangle_{\scriptscriptstyle\mathbb{E}}$, where $\langle \cdot, \cdot \rangle_{\scriptscriptstyle\mathbb{E}}$ is the Euclidean inner product.

\textbf{Geodesics.} The shortest path between two points on the manifold, written as $d_{\scriptscriptstyle\mathbb{L}}(\bm{x}, \bm{y}) = \frac{1}{\sqrt{\kappa}}\text{arcosh}(-\kappa\langle \bm{x}, \bm{y} \rangle_{\scriptscriptstyle\mathbb{L}})$.

\textbf{Tangent and Manifold Projection.} 
To map Euclidean features from CLIP into $\mathbb{L}^{n,\kappa}$, we employ the \textit{exponential map}~\cite{nickel2018learning} defined at the tangent space of a reference point $\bm{z}$.
To align with this geometry, the original Euclidean feature $\tilde{\bm{x}}_{\scriptscriptstyle\mathbb{E}} \in \mathbb{R}^n$ is lifted into the $(n+1)$-dimensional ambient tangent space $T_{\bm{z}}\mathbb{L}$, satisfying the orthogonality constraint with $\bm{z}$. 
Consequently, the projection onto the manifold to obtain $\bm{x} \in \mathbb{L}^{n,\kappa}$ is formulated as:
\begin{equation}
\label{eq:exp_map}
    \!\bm{x} \!=\! \exp_{\bm{z}}^\kappa (\bm{x}_{\scriptscriptstyle\mathbb{E}}\!) \!=\! \!\cosh(\sqrt{\kappa}\|\bm{x}_{\scriptscriptstyle\mathbb{E}}\|_{\scriptscriptstyle\mathbb{L}}) \bm{z} \!+\! \!\frac{\sinh(\sqrt{\kappa}\|\bm{x}_{\scriptscriptstyle\mathbb{E}}\|_{\scriptscriptstyle\mathbb{L}})}{\sqrt{\kappa}\|\bm{x}_{\scriptscriptstyle\mathbb{E}}\|_{\scriptscriptstyle\mathbb{L}}}\!\bm{x}_{\scriptscriptstyle\mathbb{E}},
\end{equation}
where $\bm{x}_{\scriptscriptstyle\mathbb{E}}$ refers to the lifted ambient vector derived from $\tilde{\bm{x}}_{\scriptscriptstyle\mathbb{E}}$.
Conversely, the \textit{logarithmic map} serves as the inverse operation, projecting a point $\bm{x}$ from the manifold back to the tangent space at $\bm{z}$ to recover the Euclidean features:
\begin{equation}
\label{eq:log_map}
    \bm{x}_{\scriptscriptstyle\mathbb{E}} = \log_{\bm{z}}^\kappa (\bm{x}) = \frac{\text{arccosh}(-\kappa \langle \bm{z}, \bm{x} \rangle_{\scriptscriptstyle\mathbb{L}})}{\sqrt{\langle \bm{z}, \bm{x} \rangle_{\scriptscriptstyle\mathbb{L}}^2 - 1/\kappa^2}} \Pi_{T_{\bm{z}}\mathbb{L}}(\bm{x}),
\end{equation}
where $\Pi_{T_{\bm{z}}\mathbb{L}}(\bm{x}) = \bm{x} + \kappa \langle \bm{z}, \bm{x} \rangle_{\scriptscriptstyle\mathbb{L}} \bm{z}$ denotes the orthogonal projection onto the tangent space.

\subsection{Constructing Centripetal Hyperbolic Space}
\label{sec:centripetal}
To address disordered cross-modality flows, we first align the latent geometry into a \textit{centripetal hierarchy} (Figure~\ref{fig:framework}a). Formally, let $\bm{x}_1$ denote the hyperbolic textual prototypes (serving as semantic roots/targets) and $\bm{x}_0$ denote the hyperbolic visual features (serving as entailment leaves/sources). This alignment resolves path entanglement by transforming arbitrary transport paths into ordered centripetal flows from $\bm{x}_0$ to $\bm{x}_1$. The exponential boundary expansion guarantees sufficient initialization spacing for visual feature $\bm{x}_0$, reducing spatial trajectory overlap during inward transport.

\noindent\textbf{Geometric Stratification.} We explicitly impose a centripetal hierarchy by initializing learnable scalars as: $ \alpha_{\text{txt}} < \alpha_{\text{img}}$ to modulate the feature norm before tangent to manifold projection. This configuration anchors text near the origin and pushes images to the boundary (\S Figure~\ref{fig:framework}a), creating an explicit geometric prior for inward transport.

\noindent\textbf{Centripetal hyperbolic alignment.}
To solidify this hierarchy, we optimize the embeddings using joint objectives that enforce both geometric entailment and semantic discrimination. 
In Figure~\ref{fig:framework}a, we employ the \textit{hyperbolic entailment loss}~\cite{desai2023hyperbolic} to enforce a partial order where the text prototype $\bm{x}_1$ (parent) spatially entails the image feature $\bm{x}_0$ (child). 
Formally, we constrain $\bm{x}_0$ to lie within the entailment cone of $\bm{x}_1$ by minimizing the angular violation:
\begin{equation}
    \mathcal{L}_{\text{entail}} = \max\left(0, \pi - \angle\bm{0}\bm{x}_1\bm{x}_0 - \omega(\bm{x}_1) \right),
\end{equation}
where the term $\pi - \angle\bm{0}\bm{x}_1\bm{x}_0$ corresponds to the exterior angle at $\bm{x}_1$, explicitly quantifying the deviation of the child $\bm{x}_0$ from the parent's radial axis. 
The term $\omega(\bm{x}_1) = \arcsin(2H / \|\bm{x}_1\|_{\scriptscriptstyle\mathbb{L}})$ represents the cone aperture, which naturally narrows as the prototype's radius increases.

Complementing this, we use a hyperbolic contrastive loss $\mathcal{L}_{\text{con}}$ to enforce semantic discrimination. It minimizes the geodesic distance between the image feature $\bm{x}_0$ and its ground-truth text prototype $\bm{x}^c_1$, while maximizing distances to other prototypes in the support set:
\begin{equation}
    \mathcal{L}_{\text{con}} = -\log \frac{\exp\left(-d_{\scriptscriptstyle\mathbb{L}}(\bm{x}_0, \bm{x}^c_1) / \tau\right)}{\sum_{k} \exp\left(-d_{\scriptscriptstyle\mathbb{L}}(\bm{x}_0, \bm{x}^{k}_1) / \tau\right)}.
\end{equation}
\subsection{Learning Path-Decoupled Flows}
\label{sec:flow_matching}

Having aligned the centripetal hierarchy, we proceed to learn the transport dynamics that guide entailment from visual leaves to semantic roots.
Unlike standard Riemannian flow matching~\cite{chenflow2024} that typically requires integrating an ordinary differential equation (ODE) over a continuous velocity field, we propose path-decoupled HFM.
Our core insight is to exploit \textit{step-wise transport}~\cite{yang2024consistency} to enable explicit geometric supervision (\S Figure~\ref{fig:framework}b). 

\noindent\textbf{Geodesic Path.}
Similar to the plain Euclidean flow matching~\cite{jiang2025exploring}, we define the ground-truth trajectory as the geodesic path on $\mathbb{L}$ connecting the source image $\bm{x}_0$ explicitly to its ground-truth class prototype $\bm{x}_1$. This strict pairing disentangles the optimization into independent source-target tasks, effectively eliminating inter-class interference~\cite{jiang2025exploring}.

\noindent\textbf{Tangent Velocity Alignment.}
To learn the transport dynamics, we sample paired states $(\bm{x}_t, \bm{x}_{t+\delta})$ from the ground-truth geodesic path. Specifically, given a random time $t$ and a step size $\delta$, we aim to predict the update that transports the feature from $\bm{x}_t$ to $\bm{x}_{t+\delta}$.
Our network $\mathcal{F}_\theta$ predicts an ambient vector at state $\bm{x}_t$, which is explicitly projected onto the local tangent space to ensure geometric validity:
\begin{equation}
    \bm{v}_t = \Pi_{T_{\bm{x}_t}\mathbb{L}}\left( \mathcal{F}_\theta(\bm{x}_t, t) \right),
\end{equation}
where $\Pi_{T_{\bm{x}_t}\mathbb{L}}$ denotes the orthogonal projection onto the tangent space at $\bm{x}_t$. 
We then model the discrete evolution by mapping this tangent velocity back onto the manifold via the exponential map~\cite{chenflow2024}:
\begin{equation}
    \hat{\bm{x}}_{t+\delta} = \exp_{\bm{x}_t}^\kappa(\delta \cdot \bm{v}_t).
\end{equation}
This parameterization ensures that the predicted next state $\hat{\bm{x}}_{t+\delta}$ strictly resides on the hyperbolic manifold $\mathbb{L}$.

\noindent\textbf{Path-decoupled Objective.}
We optimize the network using a path-decoupled objective that enforces both step-wise trajectory consistency and semantic separation. Crucially, both losses operate on the predicted future state $\hat{\bm{x}}_{t+\delta}$, enabling direct rectification of the flow.

\noindent\textit{1) Step-wise Consistency Loss:} 
To ensure the learned flow precisely follows the geodesic path, we minimize the squared Riemannian distance between the predicted state $\hat{\bm{x}}_{t+\delta}$ and the ground-truth target $\bm{x}_{t+\delta}$:
\begin{equation}
    \mathcal{L}_{\text{step}} = \| d_{\scriptscriptstyle\mathbb{L}}(\hat{\bm{x}}_{t+\delta}, \bm{x}_{t+\delta}) \|^2.
\end{equation}
\noindent\textit{2) Inter-Class Decoupling Loss:} 
To prevent the flow from drifting into the attraction basins of incorrect classes, we impose a ``semantic guardrail'' via a dynamic hyperbolic contrastive loss. Unlike the static alignment, this objective forces the predicted intermediate state $\hat{\bm{x}}_{t+\delta}$ to maximize its similarity with the corresponding prototype $\bm{x}_1^{c}$ while repelling all negative prototypes:
\begin{equation}
    \mathcal{L}_{\text{icd}} = -\log \frac{\exp\left(-d_{\scriptscriptstyle\mathbb{L}}(\hat{\bm{x}}_{t+\delta}, \bm{x}_1^{c}) / \tau\right)}{\sum_{k} \exp\left(-d_{\scriptscriptstyle\mathbb{L}}(\hat{\bm{x}}_{t+\delta}, \bm{x}_1^{k}) / \tau\right)}.
\end{equation}
The total objective is $\mathcal{L} = \mathcal{L}_{\text{step}} + \lambda \mathcal{L}_{\text{icd}}$. The overall training procedure is summarized in Algorithm~\ref{alg:training}. By penalizing semantic deviations at each intermediate step, this mechanism strictly confines trajectories to class-specific corridors, thereby eliminating the path entanglement issue.

\begin{algorithm}[!t]
\caption{Training Path-Decoupled Flows}
\label{alg:training}
\begin{algorithmic}[1]
\STATE \textbf{Input:} paired hyperbolic image features and text features $\mathcal{D} = \{(\bm{x}_0, \bm{x}_1)\}$, stepsize $\delta$, weight $\lambda$.
\STATE \textbf{Initialize:} flow network parameters $\theta$.
\REPEAT
    \STATE $t \sim \mathcal{U}([0, 1])$,~$(\bm{x}_0, \bm{x}_1) \sim \mathcal{D}$
    \vspace{0.1em}
    \STATE $\bm{x}_t \leftarrow \exp_{\bm{x}_0}^\kappa(t \cdot \log_{\bm{x}_0}^\kappa(\bm{x}_1))$
    \vspace{0.1em}
    \STATE $\bm{x}_{t+\delta} \leftarrow \exp_{\bm{x}_0}^\kappa((t+\delta) \cdot \log_{\bm{x}_0}^\kappa(\bm{x}_1))$
    \vspace{0.1em}
    \STATE $\bm{v}_t \leftarrow \Pi_{T_{\bm{x}_t}\mathbb{L}}(\mathcal{F}_\theta(\bm{x}_t, t))$
    \vspace{0.1em}
    \STATE $\hat{\bm{x}}_{t+\delta} \leftarrow \exp_{\bm{x}_t}^\kappa(\delta \cdot \bm{v}_t)$
    \vspace{0.1em}
    \STATE $\mathcal{L} = \| d_{\scriptscriptstyle\mathbb{L}}(\hat{\bm{x}}_{t+\delta}, \bm{x}_{t+\delta}) \|^2 + \lambda \mathcal{L}_{\text{icd}}$ 
    \vspace{0.1em}
    \STATE Update $\theta$ using gradient descent to minimize $\mathcal{L}$
\UNTIL converged
\end{algorithmic}
\end{algorithm}
\begin{algorithm}[!t]
\caption{Inference with Diameter-based Stopping}
\label{alg:inference}
\begin{algorithmic}[1]
\STATE \textbf{Input:} image feature $\bm{x}_0$, text prototypes $\{\bm{x}_1^{c}\}$, trained flow network $\mathcal{F}_\theta$, stepsize $\delta$, class number $N$.
\STATE \textbf{Initialize:} $t \leftarrow 0$, $d_{\text{txt}} = \max_{i, j} d_{\scriptscriptstyle\mathbb{L}}(\bm{x}_1^{i}, \bm{x}_1^{j})$.
\WHILE{$t < 1$}
    \STATE $\bm{v}_t \leftarrow \Pi_{T_{\hat{\bm{x}}_t}\mathbb{L}}(\mathcal{F}_\theta(\hat{\bm{x}}_t, t))$
    \vspace{0.1em}
    \STATE $\hat{\bm{x}}_{t+\delta} \leftarrow \exp_{\hat{\bm{x}}_t}^\kappa(\delta \cdot \bm{v}_t)$
    \vspace{0.1em}
    \IF{$\min_c d_{\scriptscriptstyle\mathbb{L}}(\hat{\bm{x}}_{t+\delta}, \bm{x}_1^c) \leq \phi(N)\cdot d_{\text{txt}}$}
        \vspace{0.1em}
         \STATE $t^* \leftarrow t+\delta$; \textbf{break}
    \ENDIF
        \vspace{0.1em}
    \STATE $t \leftarrow t + \delta$
\ENDWHILE
\STATE \textbf{return} $\hat{y} = \operatorname*{arg\,min}_{c} \sum_{t=0}^{t^*} d_{\scriptscriptstyle\mathbb{L}}(\hat{\bm{x}}_t, \bm{x}_1^{c})$
\end{algorithmic}
\end{algorithm}
\subsection{Inference with Diameter-based Stopping}
\label{sec:inference}

During inference, we transport the visual features of test images along the predicted geodesic flow. To prevent overshooting at the crowded hyperbolic boundary, we further propose an adaptive strategy comprising two steps:

\begin{table*}[t]
\renewcommand{\arraystretch}{1.0} 
\centering
\setlength\tabcolsep{3.5pt}
\caption{\textbf{Quantitative results.} Performance comparison (\S\ref{sec:exp_sota}) of the state-of-the-art approaches. Based on CLIP-LoRA, we add HFM to further improve the performance. Top-1 accuracy averaged over 3 random seeds is reported. The highest value of each dataset is bolded.}
\label{tab:sota}
\scalebox{0.9}{
\begin{tabular}{|cr||cccccL|ccccccL|}
\hlinewd{1.1pt}
\rowcolor{mygray}
 & &
\multicolumn{6}{c|}{\textbf{Difficult}} & \multicolumn{7}{c|}{\textbf{Easy}}\\
\rowcolor{mygray}
\multirow{-2}{*}{\textbf{Shots}} & \multirow{-2}{*}{\textbf{Method}}&\textbf{Aircraft} & \textbf{SAT} & \textbf{DTD} & \textbf{SUN} & \textbf{UCF} & \textbf{Avg}& \textbf{Cars} & \textbf{Net} & \textbf{Flowers} & \textbf{Food} & \textbf{Pets} & \textbf{Caltech} & \textbf{Avg} \\
\hline
\hline
\textbf{0} & CLIP~\citeyear{radford2021learning} & 24.8 & 47.8 & 43.8 & 62.5 & 66.7&47.6 & 65.5 & 66.7 & 67.4 & 85.3 & 89.1 & 92.9 & 77.7 \\
\hline
\multirow{11}{*}{\textbf{1}}  & CoOp~\citeyear{zhou2022learning} & 20.8 & 56.4 & 50.1 & 67.0 & 71.2&53.1 & 67.5 & 65.7 & 78.3 & 84.3 & 90.2 & 92.5 & 79.8 \\
 & CoCoOp~\citeyear{zhou2022conditional} & 28.1 & 55.4 & 52.6 & 68.7 & 70.4 &55.0& 67.6 & 69.4 & 73.4 & 84.9 & 91.9 & 94.1 & 80.2\\
 & TIP-Adapter~\citeyear{zhang2022tip} & 28.8 & 67.8 & 51.6 & 67.2 & 73.4& 57.8& 67.1 & 69.4 & 83.8 & 85.8 & 90.6 & 94.0 & 81.8 \\
 & CLIP-Adapter~\citeyear{gao2024clip} & 25.2 & 49.3 & 44.2 & 65.4 & 66.9& 50.2& 65.7 & 67.9 & 71.3 & 86.1 & 89.0 & 92.0 & 78.7 \\
 & PLOT++~\citeyear{chen2022plot} & 28.6 & 65.4 & 54.6 & 66.8 & 74.3 &58.0& 68.8 & 66.5 & 80.5 & 86.2 & 91.9 & 94.3 & 81.4 \\
 & KgCoOp~\citeyear{yao2023visual} & 26.8 & 61.9 & 52.7 & 68.4 & 72.8 &56.5& 66.7 & 68.9 & 74.7 & \textbf{86.4} & \textbf{92.1} & 94.2 & 80.5 \\
 & ProGrad~\citeyear{zhu2023prompt} & 28.9 & 57.0 & 52.8 & 67.0 & 73.3&55.8 & 68.2 & 67.0 & 80.9 & 84.9 & 91.4 & 93.5 & 81.0 \\
 & CLIP-LoRA~\citeyear{zanella2024low} & 28.0 & 71.9 & 54.1 & 70.3 & 75.4& 59.9& 69.4 & \textbf{70.3} & 81.4 & 85.1 & 91.9 & 93.8 & 82.0 \\
 & ~+FMA~\citeyear{jiang2025exploring} & 28.3	&73.0&	55.1	& 70.6	&75.9& 60.6\gain{\textbf{$+$0.7}}&	69.8 &	70.2	& 
 84.9	&85.2	& \textbf{92.1}&	94.5 & 82.8\gain{\textbf{$+$0.8}} \\  
   &  ~+\textbf{HFM (Ours)} & \textbf{32.2} & \textbf{81.0} & \textbf{58.6} & \textbf{71.4} & \textbf{77.4} & \textbf{64.1}\gain{\textbf{$+$4.2}} & \textbf{70.3} & 70.0 & \textbf{86.2} & 85.4 & 92.0 & \textbf{95.2} & \textbf{83.2}\gain{\textbf{$+$1.2}} \\
  
 \hline
 \hline
\multirow{11}{*}{\textbf{4}} & CoOp~\citeyear{zhou2022learning} & 30.9 & 69.7 & 59.5 & 69.7 & 77.6&61.5 & 74.4 & 68.8 & 92.2 & 84.5 & 92.5 & 94.5 & 84.5 \\
 & CoCoOp~\citeyear{zhou2022conditional} & 30.6 & 61.7 & 55.7 & 70.4 & 75.3& 58.7& 69.5 & 70.6 & 81.5 & 86.3 & 92.7 & 94.8 & 82.6 \\
 & TIP-Adapter~\citeyear{zhang2022tip} & 35.7 & 76.8 & 59.8 & 70.8 & 78.1 &64.2& 74.1 & 70.7 & 92.1 & 86.5 & 91.9 & 94.8 & 85.0 \\
 & CLIP-Adapter~\citeyear{gao2024clip} & 27.9 & 51.2 & 46.1 & 68.0 & 70.6 &52.8& 67.5 & 68.6 & 73.1 & 86.5 & 90.8 & 94.0 & 80.1 \\
 & PLOT++~\citeyear{chen2022plot} & 35.3 & 83.2 & 62.4 & 71.7 & 79.8 &66.5& 76.3 & 70.4 & 92.9 & 86.5 & 92.7 & 95.1 & 85.6 \\
 & KgCoOp~\citeyear{yao2023visual} & 32.2 & 71.8 & 58.7 & 71.5 & 77.6 &62.4& 69.5 & 69.9 & 87.0 & \textbf{86.9} & 92.6 & 95.0 & 83.5 \\
 & ProGrad~\citeyear{gao2024clip} & 34.1 & 69.6 & 59.7 & 71.7 & 77.9 &62.6& 75.0 & 70.2 & 91.1 & 85.4 & 92.1 & 94.4 & 84.7 \\
 & CLIP-LoRA~\citeyear{gao2024clip} & 38.8& 	83.5& 	64.0&	72.8& 	81.1	& 68.0&77.4& 	71.4& 	92.9	& 82.6& 	90.6& 	95.0 & 85.0 \\
 &  ~+FMA~\citeyear{jiang2025exploring} & {40.3}	&{85.0}	&{67.0}&	{73.7}&	{82.4}& {69.7}\gain{\textbf{$+$1.7}}&	{78.9}&	{72.0}	&{95.0}	&{83.2}&	90.8&	{95.8}& {86.0}\gain{\textbf{$+$1.0}} \\ 
  &  ~+\textbf{HFM (Ours)} & 
  \textbf{43.5} & \textbf{90.3} & \textbf{68.6} & \textbf{75.5} & \textbf{83.6} & \textbf{72.3}\gain{\textbf{$+$4.3}} & \textbf{80.1} & \textbf{72.1} & \textbf{96.2} & \textbf{86.9} & \textbf{93.2} & \textbf{96.2} & \textbf{87.4}\gain{\textbf{$+$2.4}} \\
 \hline
 \hline
\multirow{11}{*}{\textbf{16}} & CoOp~\citeyear{zhou2022learning} & 43.3	&86.0	&70.0	&74.9&	83.1&71.4&	83.1&	71.4	&97.2&	84.4&	91.1&	95.5 & 87.1\\
 & CoCoOp~\citeyear{zhou2022conditional} & 33.8	&75.5&	65.8&	72.8&	76.0&64.8	&72.4	&71.1&	87.1&	87.4&	93.2&	95.2 & 84.4 \\
 & TIP-Adapter~\citeyear{zhang2022tip} & 44.6 & 85.9 & 70.8 & 76.0 & 83.9 &72.2& 82.3 & 73.4 & 96.2 & 86.8 & 92.6 & 95.7 & 87.8 \\
 & CLIP-Adapter~\citeyear{gao2024clip} & 34.2 & 71.4 & 59.4 & 74.2 & 80.2& 63.9& 74.0 & 69.8 & 92.9 & 87.1 & 92.3 & 94.9 & 85.2 \\
 & PLOT++~\citeyear{chen2022plot} & 46.7 & 92.0 & 71.4 & 76.0 & 85.3& 74.3& 84.6 & 72.6 & 97.6 & 87.1 & 93.6 & 96.0 & 88.6 \\
 & KgCoOp~\citeyear{yao2023visual} & 36.5 & 76.2 & 68.7 & 73.3 & 81.7&67.3 & 74.8 & 70.4 & 93.4 & \textbf{87.2} & 93.2 & 95.2 & 85.7 \\
 & ProGrad~\citeyear{gao2024clip} & 43.0 & 83.6 & 68.8 & 75.1 & 82.7&70.6 & 82.9 & 72.1 & 96.6 & 85.8 & 92.8 & 95.9 & 87.7 \\
 & CLIP-LoRA~\citeyear{gao2024clip} & 54.7&	90.7	&73.0	&76.0	&86.2	&76.1 &86.0 &	73.4&	97.9	&84.2	&91.6	&96.1& 88.2 \\
 &  ~+FMA~\citeyear{jiang2025exploring} & {57.8}&	91.0	&{75.4}	&{77.2}&	{87.1}&{77.7}\gain{\textbf{$+$1.6}}&	{87.7}	&{73.5}	&\textbf{99.1}	&85.1	&91.6	&{96.5}& {88.9}\gain{\textbf{$+$0.7}} \\ 
  &  ~+\textbf{HFM (ours)} & 
  \textbf{62.1} & \textbf{94.3} & \textbf{76.0} & \textbf{77.5} & \textbf{88.9} & \textbf{79.8}\gain{\textbf{$+$3.7}} & \textbf{88.6} & \textbf{73.6} & 98.7 & \textbf{87.2} & \textbf{94.1} & \textbf{96.8} & \textbf{89.8}\gain{\textbf{$+$1.6}}
  \\
 \hline
\end{tabular}
}
\vspace{-0.5em}
\end{table*}
\noindent\textbf{Riemannian Euler Integration.}
We solve the transport ODE using a discrete Euler method~\cite{lipman2022flow}. Specifically, starting from the hyperbolic visual feature $\hat{\bm{x}}_0$, we iteratively update the state by projecting the predicted tangent velocity back onto the manifold:
\begin{equation}
    \hat{\bm{x}}_{t+\delta} = \exp_{\hat{\bm{x}}_t}^\kappa \left(\delta \cdot \Pi_{T_{\hat{\bm{x}}_t}\mathbb{L}}(\mathcal{F}_\theta(\hat{\bm{x}}_t, t))\right).
\end{equation}
\noindent\textbf{Diameter-based Stopping.}
We define the semantic diameter $d_{\text{txt}}$ as the maximum pairwise geodesic distance among all target prototypes $\{\bm{x}_1^{c}\}$.
During inference, transport terminates at step $t^*$ once the geodesic distance to the nearest prototype falls within a density-adaptive threshold:
\begin{equation}
    \min\nolimits_{c} d_{\scriptscriptstyle\mathbb{L}}(\hat{\bm{x}}_{t^*}, \bm{x}_1^{c}) \leq \phi(N) \cdot d_{\text{txt}}.
\end{equation}
Here, $d_{\text{txt}} = \max_{i, j} d_{\scriptscriptstyle\mathbb{L}}(\bm{x}_1^{i}, \bm{x}_1^{j})$ represents the intrinsic semantic scale, while $\phi(N) = 0.5 \log_{10}(N)$ compensates for manifold crowding as the class cardinality $N$ grows.

To mitigate local fluctuations, instead of relying solely on the final state, we ensemble the class probabilities across all valid steps up to $t^*$ to determine the final prediction.
The entire inference process is presented in Algorithm~\ref{alg:inference}.

\section{Experiments}
\vspace{0.5em}
\noindent\textbf{Datasets.} We evaluated HFM on standard few-shot image classification. Particularly, we conducted experiments on 11 benchmarks, including Aircraft~\citep{maji2013fine}, EuroSAT~\citep{helber2019eurosat}, DTD~\citep{cimpoi2014describing}, SUN397~\citep{xiao2010sun}, UCF101~\citep{soomro2012ucf101}, StanfordCars~\citep{krause20133d}, ImageNet~\citep{deng2009imagenet}, Flowers102~\citep{nilsback2008automated}, Food101~\citep{bossard2014food}, OxfordPets~\citep{parkhi2012cats}, and Caltech101~\citep{fei2004learning}. 
Following prior works~\cite{jiang2025exploring}, we partitioned them into two subsets: the first five datasets form the \textit{difficult} group, while the remaining six constitute the \textit{easy} group.
For each dataset, we adopted the standard train/validation/test splits. Under the $K$-shot setting, we constructed the training set by uniformly sampling $K$ labeled images per class, with the rest used for validation and testing according to the protocol. 

\begin{table*}[t]
\centering
\caption{\textbf{Ablation Study.} Analysis of component effectiveness (\S\ref{sec:exp_component}) under 4/16-shot settings. \textbf{CHA}: \underline{C}entripetal \underline{H}yperbolic \underline{A}lignment to construct the centripetal hierarchy, \textbf{PO}: flow matching with the \underline{P}ath-decoupled \underline{O}bjective, \textbf{DS}: inference with \underline{D}iameter-based \underline{S}topping.}
\label{tab:abla_component}
\renewcommand{\arraystretch}{1.0} 
\setlength\tabcolsep{3pt} 

\scalebox{0.9}{
\begin{tabular}{|c|ccc||cccccL|ccccccL|}
\hlinewd{1.1pt}
\rowcolor{mygray}
 & \multicolumn{3}{c||}{\textbf{Components}} & \multicolumn{6}{c|}{\textbf{Difficult Datasets}} & \multicolumn{7}{c|}{\textbf{Easy Datasets}} \\
\rowcolor{mygray}
\multirow{-2}{*}{\textbf{Shots}} & \textbf{CHA} & \textbf{PO} & \textbf{DS} & \textbf{FGVC} & \textbf{EuroSAT} & \textbf{DTD} & \textbf{SUN} & \textbf{UCF} & \textbf{Avg} & \textbf{Cars} & \textbf{Net} & \textbf{Flowers} & \textbf{Food} & \textbf{Pets} & \textbf{Caltech} & \textbf{Avg} \\
\hline
\hline
\multirow{4}{*}{\textbf{4}} 
& \textcolor{mygray}{\ding{55}} & \textcolor{mygray}{\ding{55}} & \textcolor{mygray}{\ding{55}} & 38.8 & 83.5 & 64.0 & 72.8 & 81.1 & 68.0 & 77.4 & 71.4 & 92.9 & 82.6 & 90.6 & 95.0 & 85.0 \\
& \ding{51} & \textcolor{mygray}{\ding{55}} & \textcolor{mygray}{\ding{55}} & 38.9 & 88.4 & 65.9 & 74.0 & 81.8 & 69.8 & 77.6 & 71.4 & 94.4 & 86.7 & 92.6 & 95.7 & 86.4 \\
& \ding{51} & \ding{51} & \textcolor{mygray}{\ding{55}} & 43.5 & 88.6 & 68.0 & 75.4 & 83.6 & 71.8 & 80.5 & 71.9 & 96.4 & 86.8 & 92.7 & 95.8 & 87.3 \\
& \ding{51} & \ding{51} & \ding{51} & 43.5 & 90.3 & 68.6 & 75.5 & 83.6 & 72.3 & 80.1 & 72.1 & 96.2 & 86.9 & 93.2 & 96.2 & 87.4 \\
\hline
\hline
\multirow{4}{*}{\textbf{16}} 
& \textcolor{mygray}{\ding{55}} & \textcolor{mygray}{\ding{55}} & \textcolor{mygray}{\ding{55}} & 54.7 & 90.7 & 73.0 & 76.0 & 86.2 & 76.1 & 86.0 & 73.4 & 97.9 & 84.2 & 91.6 & 96.1 & 88.2 \\
& \ding{51} & \textcolor{mygray}{\ding{55}} & \textcolor{mygray}{\ding{55}} & 56.9 & 93.2 & 74.1 & 76.9 & 87.6 & 77.7 & 86.6 & 73.6 & 98.4 & 87.3 & 93.5 & 96.4 & 89.3 \\
& \ding{51} & \ding{51} & \textcolor{mygray}{\ding{55}} & 61.4 & 93.2 & 76.4 & 77.5 & 87.3 & 79.2 & 88.5 & 74.3 & 98.5 & 87.4 & 93.7 & 96.5 & 89.8 \\
& \ding{51} & \ding{51} & \ding{51} & 62.1 & 94.3 & 76.0 & 77.5 & 88.9 & 79.8 & 88.6 & 73.6 & 98.7 & 87.2 & 94.1 & 96.8 & 89.8 \\
\hline
\end{tabular}%
}
\end{table*}
\begin{table*}[t]
\renewcommand{\arraystretch}{1.0} 
\centering
\setlength\tabcolsep{3.5pt}
\caption{\textbf{ Ablation Study.} Analysis of model-agnostic capability (\S\ref{sec:exp_peft}) across various PEFT approaches under the 16-shot setting.}
\label{tab:abla_peft}
\scalebox{0.9}{
\begin{tabular}{|r||cccccL|ccccccL|}
\hlinewd{1.1pt}
\rowcolor{mygray}
 &
\multicolumn{6}{c|}{\textbf{Difficult}} & \multicolumn{7}{c|}{\textbf{Easy}}\\
\rowcolor{mygray}
\multirow{-2}{*}{\textbf{Method}}&\textbf{Aircraft} & \textbf{SAT} & \textbf{DTD} & \textbf{SUN} & \textbf{UCF} & \textbf{Avg}& \textbf{Cars} & \textbf{Net} & \textbf{Flowers} & \textbf{Food} & \textbf{Pets} & \textbf{Caltech} & \textbf{Avg} \\
\hline
\hline
CoOp~\citeyear{zhou2022learning} & 43.3	&86.0	&70.0	&74.9&	83.1&71.4&	83.1&	71.4	&97.2&	84.4&	91.1&	95.5 & 87.1\\
~+FMA~\citeyear{jiang2025exploring} & 47.6 & 88.1 & 73.1 & 75.9 & 84.4 & 73.8\gain{\textbf{$+$2.4}} & 85.4 & 72.5 & 98.2 & 85.0 & 91.4 & 95.7 & 88.0\gain{\textbf{$+$0.9}} \\
~+\textbf{HFM (ours)} & 49.2 & 88.9 & 75.1 & 76.4 & 84.7 & 74.9\gain{\textbf{$+$3.5}} & 84.9 & 72.2 & 98.2 & 86.6 & 93.8 & 96.3 & 88.7\gain{\textbf{$+$1.6}} \\
\hline
 CoCoOp~\citeyear{zhou2022conditional} & 33.8	&75.5&	65.8&	72.8&	76.0&64.8	&72.4	&71.1&	87.1&	87.4&	93.2&	95.2 & 84.4 \\
~+FMA~\citeyear{jiang2025exploring} & 36.9 & 86.9 & 71.9 & 73.4 & 80.3 & 69.9\gain{\textbf{$+$5.1}} & 73.5 & 71.9 & 94.5 & 87.8 & 93.4 & 95.6 & 86.1\gain{\textbf{$+$1.7}} \\
~+\textbf{HFM (ours)} & 43.7 & 89.3 & 72.6 & 75.6 & 83.9 & 73.0\gain{\textbf{$+$8.2}}  & 79.3 & 71.9 & 97.9 & 87.1 & 93.9 & 96.3 & 87.7\gain{\textbf{$+$3.3}}  \\
 \hline
 CLIP-Adapter~\citeyear{gao2024clip} & 33.8 & 70.4 & 59.3 & 74.3 & 80.1 & 63.6 & 74.2 & 71.6 & 93.6 & 87.1 & 92.4 & 94.9 & 85.6 \\
~+FMA~\citeyear{jiang2025exploring} & 35.8 & 85.6 & 69.2 & 74.4 & 81.5 & 69.3\gain{\textbf{$+$5.7}} & 74.7 & 71.3 & 95.6 & 87.2 & 92.9 & 96.0 & 86.3\gain{\textbf{$+$0.7}} \\
~+\textbf{HFM (ours)} & 43.8 & 85.7 & 71.4 & 76.1 & 84.6 & 72.3\gain{\textbf{$+$8.7}} & 80.6 & 72.1 & 97.2 & 87.0 & 93.3 & 96.2 & 87.7\gain{\textbf{$+$2.1}}\\
 \hline
 CLIP-LoRA~\citeyear{gao2024clip} & 54.7&	90.7	&73.0	&76.0	&86.2	&76.1 &86.0 &	73.4&	97.9	&84.2	&91.6	&96.1& 88.2 \\
 ~+FMA~\citeyear{jiang2025exploring} & {57.8}&	91.0	&{75.4}	&{77.2}&	{87.1}&{77.7}\gain{\textbf{$+$1.6}}&	{87.7}	&{73.5}	& {99.1}	&85.1	&91.6	&{96.5}& {88.9}\gain{\textbf{$+$0.7}} \\
~+\textbf{HFM (ours)} & 
  {62.1} & {94.3} & {76.0} & {77.5} & {88.9} & {79.8}\gain{\textbf{$+$3.7}} & {88.6} & {73.6} & 98.7 & {87.2} & {94.1} & {96.8} & {89.8}\gain{\textbf{$+$1.6}}
  \\
 \hline
\end{tabular}
}
 \vspace{-0.5em}
\end{table*}

\noindent\textbf{Implementation Details.}
We implemented HFM on top of CLIP-LoRA~\citep{zanella2024low}. Specifically, we first utilized the trained CLIP-LoRA model after centripetal hyperbolic alignment to extract image and text features, then trained a velocity network on these features following the HFM framework.
Similar to FMA~\cite{jiang2025exploring}, we adopted the lightweight architecture from MAR~\cite{li2024autoregressive} as our flow-matching network, implemented as a deep residual MLP with timestep conditioning. As for Lorentz model, we initialized the manifold curvature $\kappa=1.0$ and $\alpha_{\text{txt}} = 0.5 \cdot\alpha_{\text{img}}$. Following~\cite{desai2023hyperbolic}, we set a constant $H$=0.1. The balance weight $\lambda$ for the inter-class decoupling loss was set to 0.1. We optimized HFM with AdamW~\citep{loshchilov2017decoupled} using a learning rate of $2\times10^{-4}$ and a cosine annealing schedule. 

\subsection{Quantitative Comparison}
\vspace{0.2em}
\label{sec:exp_sota}
\noindent\textbf{Settings.} We compared HFM with recent few-shot classification methods including CLIP~\citep{radford2021learning}, CoOp~\citep{zhou2022learning}, CoCoOp~\citep{zhou2022conditional}, TIP-Adapter~\citep{zhang2022tip}, CLIP-Adapter~\citep{gao2024clip}, PLOT++~\citep{chen2022plot}, KgCoOp~\citep{yao2023visual}, ProGrad~\citep{zhu2023prompt}, CLIP-LoRA~\citep{zanella2024low} and FMA~\cite{jiang2025exploring}. All methods use CLIP ViT-B/16~\citep{radford2021learning} as the backbone.


\noindent\textbf{Results.} As seen from Table~\ref{tab:sota}, HFM consistently outperforms SOTA baselines across all settings. 
Crucially, HFM surpasses the Euclidean counterpart FMA~\cite{jiang2025exploring}, verifying the efficacy of hyperbolic geometry in resolving path entanglement. On the difficult benchmarks, HFM achieves \textbf{64.1\%} (1-shot) and \textbf{79.8\%} (16-shot), exceeding FMA by \textbf{3.5\%} and \textbf{2.1\%}, respectively.
This advantage is most pronounced on structurally complex datasets like EuroSAT and DTD, where HFM outperforms FMA by \textbf{8.0\%} and \textbf{3.5\%} under the 1-shot setting.
Furthermore, compared to the strong CLIP-LoRA~\citep{zanella2024low} baseline, HFM yields consistent gains of \textbf{3.7\%}--\textbf{4.3\%} on difficult datasets, demonstrating the superiority of continuous flow adaptation over static parameter tuning.

\subsection{Diagnostic Experiments}

\begin{table*}[t]
\renewcommand{\arraystretch}{1.1} 
\centering
\setlength\tabcolsep{3pt}
\caption{\small \textbf{Ablation study.} Performance comparison (\S\ref{sec:exp_backbone}) between CLIP-LoRA and our HFM on different backbones (ViT-B32, ViT-B16, and ViT-L14) with 4 and 16 shots. Top-1 accuracy is reported. The highest value of each setting is bolded.}
\label{tab:backbone}
\scalebox{0.85}{
\begin{tabular}{|c|cr||cccccL|ccccccL|}
\hlinewd{1.1pt}
\rowcolor{mygray}
 & & & \multicolumn{6}{c|}{\textbf{Difficult}} & \multicolumn{7}{c|}{\textbf{Easy}}\\
\rowcolor{mygray}
\multirow{-2}{*}{\textbf{Shots}} & \multirow{-2}{*}{\textbf{Backbone}} & \multirow{-2}{*}{\textbf{Method}}&\textbf{Aircraft} & \textbf{SAT} & \textbf{DTD} & \textbf{SUN} & \textbf{UCF} & \textbf{Avg}& \textbf{Cars} & \textbf{Net} & \textbf{Flowers} & \textbf{Food} & \textbf{Pets} & \textbf{Caltech} & \textbf{Avg} \\
\hline
\hline
\multirow{6}{*}{\textbf{4}} 
& \multirow{2}{*}{ViT-B32} & CLIP-LoRA~\citeyear{gao2024clip} & 27.7 & 85.6 & 60.3 & 70.3 & 76.5 & 64.1 & 68.3 & 66.5 & 90.1 & 75.6 & 86.3 & 94.3 & 80.2 \\
& & ~+\textbf{HFM (ours)} & \textbf{33.9} & \textbf{90.1} & \textbf{64.5} & \textbf{72.7} & \textbf{80.1} & \textbf{68.3}\gain{\textbf{$+$4.2}} & \textbf{71.6} & \textbf{66.7} & \textbf{93.1} & \textbf{80.7} & \textbf{90.6} & \textbf{95.4} & \textbf{83.0}\gain{\textbf{$+$2.8}} \\
\cline{2-16}
& \multirow{2}{*}{ViT-B16} & CLIP-LoRA~\citeyear{gao2024clip} & 38.8 & 83.5 & 64.0 & 72.8 & 81.1 & 68.0 & 77.4 & 71.4 & 92.9 & 82.6 & 90.6 & 95.0 & 85.0 \\
& & ~+\textbf{HFM (ours)} & \textbf{43.5} & \textbf{90.3} & \textbf{68.6} & \textbf{75.5} & \textbf{83.6} & \textbf{72.3}\gain{\textbf{$+$4.3}} & \textbf{80.1} & \textbf{72.1} & \textbf{96.2} & \textbf{86.9} & \textbf{93.2} & \textbf{96.2} & \textbf{87.4}\gain{\textbf{$+$2.4}} \\
\cline{2-16}
& \multirow{2}{*}{ViT-L14} & CLIP-LoRA~\citeyear{gao2024clip} & 48.9 & 86.4 & 70.4 & 76.7 & 86.4 & 73.8 & 85.2 & 77.9 & 97.4 & 89.6 & 93.9 & 97.2 & 90.2 \\
& & ~+\textbf{HFM (ours)} & \textbf{54.8} & \textbf{88.6} & \textbf{72.2} & \textbf{78.9} & \textbf{88.3} & \textbf{76.6}\gain{\textbf{$+$2.8}} & \textbf{87.0} & \textbf{78.9} & \textbf{98.7} & \textbf{91.8} & \textbf{95.3} & \textbf{97.7} & \textbf{91.6}\gain{\textbf{$+$1.4}} \\
\hline
\hline
\multirow{6}{*}{\textbf{16}} 
& \multirow{2}{*}{ViT-B32} & CLIP-LoRA~\citeyear{gao2024clip} & 44.9 & 91.8 & 68.2 & 74.0 & 82.8 & 72.3 & 79.7 & 68.4 & 96.2 & 78.2 & 88.8 & 95.2 & 84.4 \\
& & ~+\textbf{HFM (ours)} & \textbf{50.4} & \textbf{93.1} & \textbf{71.0} & \textbf{75.0} & \textbf{85.2} & \textbf{74.9}\gain{\textbf{$+$2.6}} & \textbf{81.8} & \textbf{68.5} & \textbf{97.3} & \textbf{81.0} & \textbf{89.3} & \textbf{95.9} & \textbf{85.6}\gain{\textbf{$+$1.2}} \\
\cline{2-16}
& \multirow{2}{*}{ViT-B16} & CLIP-LoRA~\citeyear{gao2024clip} & 54.7 & 90.7 & 73.0 & 76.0 & 86.2 & 76.1 & 86.0 & 73.4 & 97.9 & 84.2 & 91.6 & 96.1 & 88.2 \\
& & ~+\textbf{HFM (ours)} & \textbf{62.1} & \textbf{94.3} & \textbf{76.0} & \textbf{77.5} & \textbf{88.9} & \textbf{79.8}\gain{\textbf{$+$3.7}} & \textbf{88.6} & \textbf{73.6} & \textbf{98.7} & \textbf{87.3} & \textbf{94.1} & \textbf{96.8} & \textbf{89.8}\gain{\textbf{$+$1.6}} \\
\cline{2-16}
& \multirow{2}{*}{ViT-L14} & CLIP-LoRA~\citeyear{gao2024clip} & 66.2 & 93.1 & 76.5 & 79.4 & 89.9 & 81.0 & 90.9 & 79.6 & 99.0 & 89.9 & 94.3 & 97.3 & 91.8 \\
& & ~+\textbf{HFM (ours)} & \textbf{68.7} & \textbf{93.4} & \textbf{80.1} & \textbf{81.6} & \textbf{91.5} & \textbf{83.1}\gain{\textbf{$+$2.1}} & \textbf{92.0} & \textbf{80.7} & \textbf{99.3} & \textbf{92.2} & \textbf{96.2} & \textbf{98.0} & \textbf{93.1}\gain{\textbf{$+$1.3}} \\
\hline
\end{tabular}
}
\end{table*}
To verify the efficacy of HFM and each module, we performed extensive diagnostic experiments.

\noindent\textbf{Key Component Analysis.}
\label{sec:exp_component}
Contributions of Centripetal Hyperbolic Alignment (CHA), flowing matching with Path-decoupled Objective (PO), and inference with Diameter-based Stopping (DS) were evaluated under both 4-shot and 16-shot settings, as summarized in Table~\ref{tab:abla_component}. The first row in each block refers to the baseline model (\ie, CLIP-LoRA~\citep{zanella2024low}) implemented in Euclidean space. Taking the 16-shot setting as an example, four crucial conclusions can be drawn.
\textbf{First}, by simply restructuring the latent space with CHA, consistent improvements are observed, with notable gains on the challenging Aircraft benchmark (from 54.7\% to \textbf{56.9}\%). This demonstrates that the concentric hierarchy, \ie, anchoring text as roots and images as leaves, provides a superior geometric initialization than the unstructured Euclidean space.
\textbf{Second}, with the guidance of the path-decoupled objective in flow matching, the model effectively learns to decouple transport trajectories, resulting in a substantial performance leap (\eg, \textbf{61.4}\% on Aircraft). This indicates that the ``semantic guardrail'' mechanism successfully reduces inter-class interference by confining features to isolated geodesic corridors.
\textbf{Third}, benefiting from the adaptive diameter-based stopping, the inference process terminates dynamically based on the intrinsic semantic scale. This yields further improvements (\eg, \textbf{62.1}\% on Aircraft), confirming that preventing over-transportation into the crowded origin is essential for preserving discriminability.
\textbf{Finally}, this progressive improvement is consistent across different data settings. For the difficult datasets, the average accuracy increases from 68.0\% to \textbf{72.3}\% in 4-shot setting, paralleling the trend in the 16-shot setting (76.1\% to \textbf{79.8}\%). This consistency verifies that HFM is robust to data scarcity and can effectively resolve path-entanglement regardless of the support set size.

\noindent\textbf{Model-Agnostic Generalization.}
\label{sec:exp_peft}
We investigated the generalizability of HFM by integrating it with various PEFT architectures, as shown in Table~\ref{tab:abla_peft}. As a plug-and-play module, HFM consistently yields superior performance when applied to CoOp~\citep{zhou2022learning}, CoCoOp~\citep{zhou2022conditional}, CLIP-Adapter~\citep{gao2024clip}, and CLIP-LoRA~\citep{zanella2024low}. 
For instance, when equipping CLIP-Adapter~\citep{gao2024clip} with HFM, the average accuracy on difficult datasets yields a remarkable surge of \textbf{8.7\%} (from 63.6\% to 72.3\%), significantly outperforming the \textbf{5.7\%} gain obtained by the Euclidean counterpart FMA~\cite{jiang2025exploring}. 
Similarly, on CoCoOp~\citep{zhou2022conditional}, HFM achieves an \textbf{8.2\%} improvement, surpassing FMA by a clear margin of \textbf{3.1\%}. 
Even on the stronger CoOp~\citep{zhou2022learning} baseline, HFM provides a consistent \textbf{3.5\%} gain on difficult benchmarks. 
These results confirm that the benefit of hyperbolic geometry is orthogonal to the choice of parameter-tuning strategy, serving as a universal geometric enhancer for resolving path-entanglement.

\begin{figure*}
    \centering
    \includegraphics[width=1\linewidth]{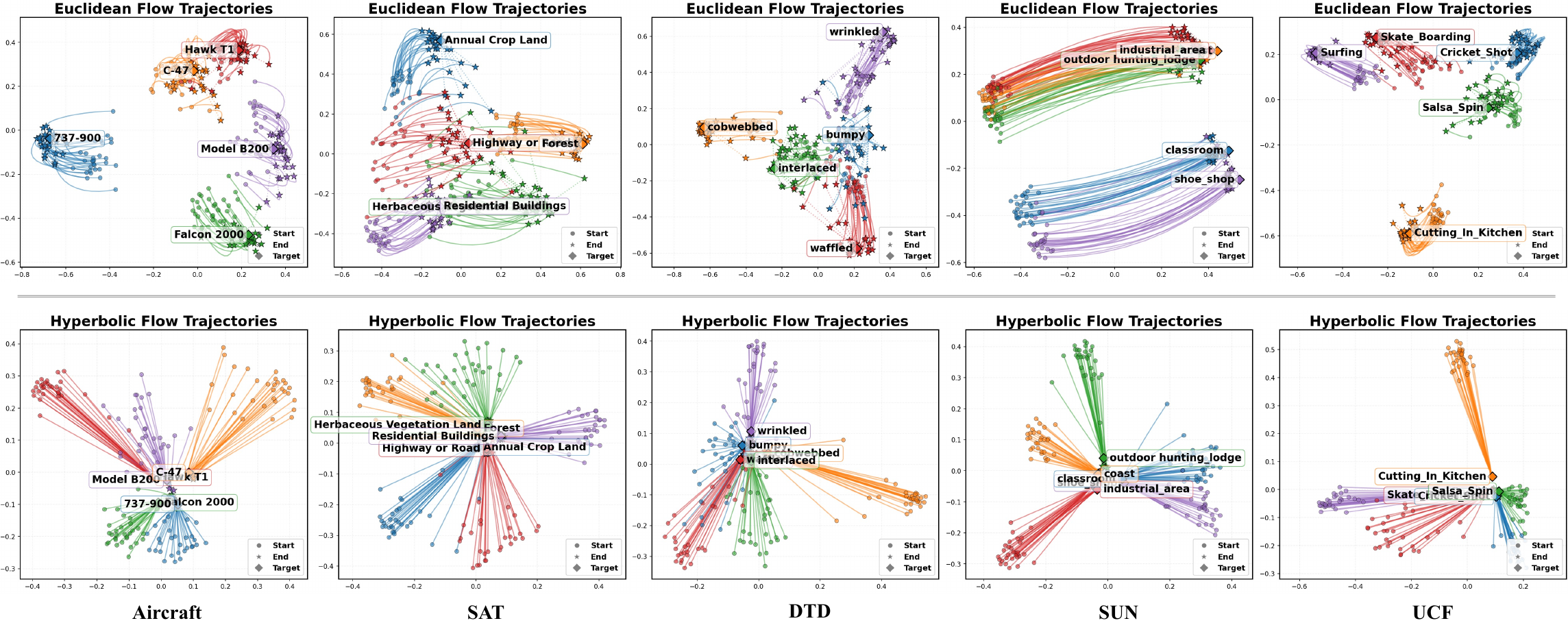}
\caption{\textbf{Qualitative Results.} Visualization of transport trajectories (\S\ref{sec:vis}).
\textbf{Top:} Euclidean flows suffer from severe \textit{path entanglement}, exhibiting chaotic and intersecting paths due to spatial crowding. 
\textbf{Bottom:} HFM achieves \textit{path decoupling} via a \textit{centripetal hierarchy}. Visual features move centripetally from the boundary to central text roots along isolated geodesic corridors.}
\label{fig:vis}
\vspace{-0.5em}
\end{figure*}

\noindent\textbf{Scalability across Backbones.}
\label{sec:exp_backbone}
Table~\ref{tab:backbone} reports the performance using different CLIP backbones (ViT-B/32, ViT-B/16, and ViT-L/14) under both 4/16-shot settings.
HFM consistently enhances the baseline CLIP-LoRA across all model sizes.
Specifically, on the difficult benchmarks, HFM achieves remarkable accuracy gains.
For instance, with the ViT-B/32 backbone, our method improves performance by \textbf{2.6\%} (72.3\% \vs 74.9\%) in the 16-shot setting.
This advantage scales effectively to the larger ViT-L/14 model, where HFM yields consistent improvements of \textbf{2.1\%} (81.0\% \vs 83.1\%) on difficult datasets, while reaching a peak average accuracy of \textbf{93.1\%} on the easy group.
Furthermore, in the data-scarce 4-shot setup, HFM boosts the widely used ViT-B/16 backbone by a substantial margin of \textbf{4.3\%} on difficult groups.
This demonstrates that our method scales effectively with model capacity and feature dimensionality, maintaining its superiority even with stronger visual encoders.

\subsection{Qualitative Comparison}
\label{sec:vis}

To intuitively understand how HFM resolves path entanglement, we visualized the transport trajectories on five difficult benchmarks using PCA. As illustrated in Figure~\ref{fig:vis}, we can observe that:
1) \noindent\textbf{Chaotic Crossovers in Euclidean Flows.}
The top row reveals that Euclidean flow matching suffers from severe \textit{path entanglement}. Due to the limited spatial capacity of flat geometry, trajectories from different classes are densely packed and frequently intersect. For instance, in the Aircraft and DTD datasets, the flow paths of neighboring categories (\eg, different textures or aircraft models) exhibit chaotic crossovers, leading to ambiguous decision boundaries and potential misclassification. 2) \noindent\textbf{Ordered Decoupling in Hyperbolic Flows.}
In contrast, the bottom row demonstrates that HFM generates highly organized, radial trajectories. Driven by our \textit{centripetal hyperbolic alignment}, the visual features are transported from the boundary inward to the central text roots along isolated geodesic corridors. This structure effectively exploits the exponential volume expansion of the Lorentz manifold to separate diverse semantic clusters. Consequently, inter-class collisions are virtually eliminated, confirming the efficacy of our path-decoupled objective in preserving semantic discriminability during the adaptation process.

\section{Conclusion}
In this paper, we identified that the polynomial volume growth of Euclidean space limits existing flow matching methods, leading to severe path entanglement in few-shot adaptation. To this end, we proposed path-decoupled Hyperbolic Flow Matching (HFM), which exploits the exponential expansion of the Lorentz manifold to decouple transport trajectories. Through centripetal hyperbolic alignment, a path-decoupled objective, and adaptive diameter-based stopping, HFM constructs isolated geodesic corridors and ensures precise inference termination. Empirical results on 11 benchmarks confirm that HFM not only establishes a new state-of-the-art but also demonstrates remarkable data efficiency in complex visual scenarios. We hope this work encourages further exploration into non-Euclidean generative dynamics for robust cross-modal understanding.

\section*{Impact Statements}
This paper presents work whose goal is to advance the field of machine learning. There are many potential societal consequences of our work, none of which we feel must be specifically highlighted here.

\bibliography{example_paper}

@String(CVPR  = {IEEE Conf. Comput. Vis. Pattern Recog.})

@String(ICLR  = {Int. Conf. Learn. Represent.})

@String(ICASSP= {ICASSP})

@String(CVPR  = {CVPR})

@String(NIPS = {NeurIPS})

@String(ICLR  = {ICLR})

@inproceedings{rombach2022high,
  title={High-resolution image synthesis with latent diffusion models},
  author={Rombach, Robin and Blattmann, Andreas and Lorenz, Dominik and Esser, Patrick and Ommer, Bj{\"o}rn},
  booktitle={Proceedings of the IEEE/CVF conference on computer vision and pattern recognition},
  pages={10684--10695},
  year={2022}
}

@inproceedings{esser2024scaling,
  title={Scaling rectified flow transformers for high-resolution image synthesis},
  author={Esser, Patrick and Kulal, Sumith and Blattmann, Andreas and Entezari, Rahim and M{\"u}ller, Jonas and Saini, Harry and Levi, Yam and Lorenz, Dominik and Sauer, Axel and Boesel, Frederic and others},
  booktitle={Forty-first international conference on machine learning},
  year={2024}
}

@inproceedings{radford2021learning,
  title={Learning transferable visual models from natural language supervision},
  author={Radford, Alec and Kim, Jong Wook and Hallacy, Chris and Ramesh, Aditya and Goh, Gabriel and Agarwal, Sandhini and Sastry, Girish and Askell, Amanda and Mishkin, Pamela and Clark, Jack and others},
  booktitle={International conference on machine learning},
  pages={8748--8763},
  year={2021},
  organization={PmLR}
}

@inproceedings{jia2021scaling,
  title={Scaling up visual and vision-language representation learning with noisy text supervision},
  author={Jia, Chao and Yang, Yinfei and Xia, Ye and Chen, Yi-Ting and Parekh, Zarana and Pham, Hieu and Le, Quoc and Sung, Yun-Hsuan and Li, Zhen and Duerig, Tom},
  booktitle={International conference on machine learning},
  pages={4904--4916},
  year={2021},
  organization={PMLR}
}

@article{zhou2022learning,
  title={Learning to prompt for vision-language models},
  author={Zhou, Kaiyang and Yang, Jingkang and Loy, Chen Change and Liu, Ziwei},
  journal={International Journal of Computer Vision},
  volume={130},
  number={9},
  pages={2337--2348},
  year={2022},
  publisher={Springer}
}

@inproceedings{zhou2022conditional,
  title={Conditional prompt learning for vision-language models},
  author={Zhou, Kaiyang and Yang, Jingkang and Loy, Chen Change and Liu, Ziwei},
  booktitle={Proceedings of the IEEE/CVF conference on computer vision and pattern recognition},
  pages={16816--16825},
  year={2022}
}

@article{gao2024clip,
  title={Clip-adapter: Better vision-language models with feature adapters},
  author={Gao, Peng and Geng, Shijie and Zhang, Renrui and Ma, Teli and Fang, Rongyao and Zhang, Yongfeng and Li, Hongsheng and Qiao, Yu},
  journal={International Journal of Computer Vision},
  volume={132},
  number={2},
  pages={581--595},
  year={2024},
  publisher={Springer}
}

@article{hu2022lora,
  title={Lora: Low-rank adaptation of large language models.},
  author={Hu, Edward J and Shen, Yelong and Wallis, Phillip and Allen-Zhu, Zeyuan and Li, Yuanzhi and Wang, Shean and Wang, Lu and Chen, Weizhu and others},
  journal={ICLR},
  volume={1},
  number={2},
  pages={3},
  year={2022}
}

@article{lipman2022flow,
  title={Flow matching for generative modeling},
  author={Lipman, Yaron and Chen, Ricky TQ and Ben-Hamu, Heli and Nickel, Maximilian and Le, Matt},
  journal={arXiv preprint arXiv:2210.02747},
  year={2022}
}

@article{liu2022flow,
  title={Flow straight and fast: Learning to generate and transfer data with rectified flow},
  author={Liu, Xingchao and Gong, Chengyue and Liu, Qiang},
  journal={arXiv preprint arXiv:2209.03003},
  year={2022}
}

@inproceedings{zanella2024low,
  title={Low-rank few-shot adaptation of vision-language models},
  author={Zanella, Maxime and Ben Ayed, Ismail},
  booktitle={Proceedings of the IEEE/CVF Conference on Computer Vision and Pattern Recognition},
  pages={1593--1603},
  year={2024}
}

@inproceedings{parkhi2012cats,
  title={Cats and dogs},
  author={Parkhi, Omkar M and Vedaldi, Andrea and Zisserman, Andrew and Jawahar, CV},
  booktitle={2012 IEEE conference on computer vision and pattern recognition},
  pages={3498--3505},
  year={2012},
  organization={IEEE}
}

@article{maji2013fine,
  title={Fine-grained visual classification of aircraft},
  author={Maji, Subhransu and Rahtu, Esa and Kannala, Juho and Blaschko, Matthew and Vedaldi, Andrea},
  journal={arXiv preprint arXiv:1306.5151},
  year={2013}
}

@inproceedings{deng2009imagenet,
  title={Imagenet: A large-scale hierarchical image database},
  author={Deng, Jia and Dong, Wei and Socher, Richard and Li, Li-Jia and Li, Kai and Fei-Fei, Li},
  booktitle={2009 IEEE conference on computer vision and pattern recognition},
  pages={248--255},
  year={2009},
  organization={Ieee}
}

@inproceedings{xiao2010sun,
  title={Sun database: Large-scale scene recognition from abbey to zoo},
  author={Xiao, Jianxiong and Hays, James and Ehinger, Krista A and Oliva, Aude and Torralba, Antonio},
  booktitle={2010 IEEE computer society conference on computer vision and pattern recognition},
  pages={3485--3492},
  year={2010},
  organization={IEEE}
}

@article{helber2019eurosat,
  title={Eurosat: A novel dataset and deep learning benchmark for land use and land cover classification},
  author={Helber, Patrick and Bischke, Benjamin and Dengel, Andreas and Borth, Damian},
  journal={IEEE Journal of Selected Topics in Applied Earth Observations and Remote Sensing},
  volume={12},
  number={7},
  pages={2217--2226},
  year={2019},
  publisher={IEEE}
}

@inproceedings{krause20133d,
  title={3d object representations for fine-grained categorization},
  author={Krause, Jonathan and Stark, Michael and Deng, Jia and Fei-Fei, Li},
  booktitle={Proceedings of the IEEE international conference on computer vision workshops},
  pages={554--561},
  year={2013}
}

@inproceedings{bossard2014food,
  title={Food-101--mining discriminative components with random forests},
  author={Bossard, Lukas and Guillaumin, Matthieu and Van Gool, Luc},
  booktitle={European conference on computer vision},
  pages={446--461},
  year={2014},
  organization={Springer}
}

@inproceedings{nilsback2008automated,
  title={Automated flower classification over a large number of classes},
  author={Nilsback, Maria-Elena and Zisserman, Andrew},
  booktitle={2008 Sixth Indian conference on computer vision, graphics \& image processing},
  pages={722--729},
  year={2008},
  organization={IEEE}
}

@inproceedings{fei2004learning,
  title={Learning generative visual models from few training examples: An incremental bayesian approach tested on 101 object categories},
  author={Fei-Fei, Li and Fergus, Rob and Perona, Pietro},
  booktitle={2004 conference on computer vision and pattern recognition workshop},
  pages={178--178},
  year={2004},
  organization={IEEE}
}

@inproceedings{cimpoi2014describing,
  title={Describing textures in the wild},
  author={Cimpoi, Mircea and Maji, Subhransu and Kokkinos, Iasonas and Mohamed, Sammy and Vedaldi, Andrea},
  booktitle={Proceedings of the IEEE conference on computer vision and pattern recognition},
  pages={3606--3613},
  year={2014}
}

@article{soomro2012ucf101,
  title={Ucf101: A dataset of 101 human actions classes from videos in the wild},
  author={Soomro, Khurram and Zamir, Amir Roshan and Shah, Mubarak},
  journal={arXiv preprint arXiv:1212.0402},
  year={2012}
}

@article{chen2022plot,
  title={Plot: Prompt learning with optimal transport for vision-language models},
  author={Chen, Guangyi and Yao, Weiran and Song, Xiangchen and Li, Xinyue and Rao, Yongming and Zhang, Kun},
  journal={arXiv preprint arXiv:2210.01253},
  year={2022}
}

@inproceedings{yao2023visual,
  title={Visual-language prompt tuning with knowledge-guided context optimization},
  author={Yao, Hantao and Zhang, Rui and Xu, Changsheng},
  booktitle={Proceedings of the IEEE/CVF conference on computer vision and pattern recognition},
  pages={6757--6767},
  year={2023}
}

@inproceedings{zhang2022tip,
  title={Tip-adapter: Training-free adaptation of clip for few-shot classification},
  author={Zhang, Renrui and Zhang, Wei and Fang, Rongyao and Gao, Peng and Li, Kunchang and Dai, Jifeng and Qiao, Yu and Li, Hongsheng},
  booktitle={European conference on computer vision},
  pages={493--510},
  year={2022},
  organization={Springer}
}

@inproceedings{zhu2023prompt,
  title={Prompt-aligned gradient for prompt tuning},
  author={Zhu, Beier and Niu, Yulei and Han, Yucheng and Wu, Yue and Zhang, Hanwang},
  booktitle={Proceedings of the IEEE/CVF international conference on computer vision},
  pages={15659--15669},
  year={2023}
}

@article{ho2020denoising,
  title={Denoising diffusion probabilistic models},
  author={Ho, Jonathan and Jain, Ajay and Abbeel, Pieter},
  journal={Advances in neural information processing systems},
  volume={33},
  pages={6840--6851},
  year={2020}
}

@article{song2020score,
  title={Score-based generative modeling through stochastic differential equations},
  author={Song, Yang and Sohl-Dickstein, Jascha and Kingma, Diederik P and Kumar, Abhishek and Ermon, Stefano and Poole, Ben},
  journal={arXiv preprint arXiv:2011.13456},
  year={2020}
}

@article{albergo2023stochastic,
  title={Stochastic interpolants: A unifying framework for flows and diffusions},
  author={Albergo, Michael S and Boffi, Nicholas M and Vanden-Eijnden, Eric},
  journal={arXiv preprint arXiv:2303.08797},
  year={2023}
}

@article{wang2020generalizing,
  title={Generalizing from a few examples: A survey on few-shot learning},
  author={Wang, Yaqing and Yao, Quanming and Kwok, James T and Ni, Lionel M},
  journal={ACM computing surveys (csur)},
  volume={53},
  number={3},
  pages={1--34},
  year={2020},
  publisher={ACM New York, NY, USA}
}

@inproceedings{lee2023read,
  title={Read-only prompt optimization for vision-language few-shot learning},
  author={Lee, Dongjun and Song, Seokwon and Suh, Jihee and Choi, Joonmyeong and Lee, Sanghyeok and Kim, Hyunwoo J},
  booktitle={Proceedings of the IEEE/CVF international conference on computer vision},
  pages={1401--1411},
  year={2023}
}

@article{loshchilov2017decoupled,
  title={Decoupled weight decay regularization},
  author={Loshchilov, Ilya and Hutter, Frank},
  journal={arXiv preprint arXiv:1711.05101},
  year={2017}
}

@article{jiang2025exploring,
  title={Exploring Cross-Modal Flows for Few-Shot Learning},
  author={Jiang, Ziqi and Wang, Yanghao and Chen, Long},
  journal={arXiv preprint arXiv:2510.14543},
  year={2025}
}

@inproceedings{dhingra2018embedding,
  title={Embedding text in hyperbolic spaces},
  author={Dhingra, Bhuwan and Shallue, Christopher and Norouzi, Mohammad and Dai, Andrew and Dahl, George},
  booktitle={Proceedings of the Twelfth Workshop on Graph-Based Methods for Natural Language Processing (TextGraphs-12)},
  pages={59--69},
  year={2018}
}

@article{zhu2020hypertext,
  title={Hypertext: Endowing fasttext with hyperbolic geometry},
  author={Zhu, Yudong and Zhou, Di and Xiao, Jinghui and Jiang, Xin and Chen, Xiao and Liu, Qun},
  journal={arXiv preprint arXiv:2010.16143},
  year={2020}
}

@inproceedings{le2019inferring,
  title={Inferring concept hierarchies from text corpora via hyperbolic embeddings},
  author={Le, Matthew and Roller, Stephen and Papaxanthos, Laetitia and Kiela, Douwe and Nickel, Maximilian},
  booktitle={Proceedings of the 57th annual meeting of the association for computational linguistics},
  pages={3231--3241},
  year={2019}
}

@article{wen2025hover,
  title={HOVER: Hyperbolic Video-Text Retrieval},
  author={Wen, Jun and Chen, Yufeng and Shi, Ruiqi and Ji, Wei and Yang, Menglin and Gao, Difei and Yuan, Junsong and Zimmermann, Roger},
  journal={IEEE Transactions on Image Processing},
  year={2025},
  publisher={IEEE}
}

@inproceedings{desai2023hyperbolic,
  title={Hyperbolic image-text representations},
  author={Desai, Karan and Nickel, Maximilian and Rajpurohit, Tanmay and Johnson, Justin and Vedantam, Shanmukha Ramakrishna},
  booktitle={International Conference on Machine Learning},
  pages={7694--7731},
  year={2023},
  organization={PMLR}
}

@inproceedings{atigh2022hyperbolic,
  title={Hyperbolic image segmentation},
  author={Atigh, Mina Ghadimi and Schoep, Julian and Acar, Erman and Van Noord, Nanne and Mettes, Pascal},
  booktitle={Proceedings of the IEEE/CVF conference on computer vision and pattern recognition},
  pages={4453--4462},
  year={2022}
}

@inproceedings{khrulkov2020hyperbolic,
  title={Hyperbolic image embeddings},
  author={Khrulkov, Valentin and Mirvakhabova, Leyla and Ustinova, Evgeniya and Oseledets, Ivan and Lempitsky, Victor},
  booktitle={Proceedings of the IEEE/CVF conference on computer vision and pattern recognition},
  pages={6418--6428},
  year={2020}
}

@inproceedings{liu2020hyperbolic,
  title={Hyperbolic visual embedding learning for zero-shot recognition},
  author={Liu, Shaoteng and Chen, Jingjing and Pan, Liangming and Ngo, Chong-Wah and Chua, Tat-Seng and Jiang, Yu-Gang},
  booktitle={Proceedings of the IEEE/CVF conference on computer vision and pattern recognition},
  pages={9273--9281},
  year={2020}
}

@inproceedings{li2023euclidean,
  title={The euclidean space is evil: Hyperbolic attribute editing for few-shot image generation},
  author={Li, Lingxiao and Zhang, Yi and Wang, Shuhui},
  booktitle={Proceedings of the IEEE/CVF international conference on computer vision},
  pages={22714--22724},
  year={2023}
}

@inproceedings{petermann2023hyperbolic,
  title={Hyperbolic audio source separation},
  author={Petermann, Darius and Wichern, Gordon and Subramanian, Aswin and Le Roux, Jonathan},
  booktitle={ICASSP 2023-2023 IEEE International Conference on Acoustics, Speech and Signal Processing (ICASSP)},
  pages={1--5},
  year={2023},
  organization={IEEE}
}

@inproceedings{hong2023hyperbolic,
  title={Hyperbolic audio-visual zero-shot learning},
  author={Hong, Jie and Hayder, Zeeshan and Han, Junlin and Fang, Pengfei and Harandi, Mehrtash and Petersson, Lars},
  booktitle={Proceedings of the IEEE/CVF international conference on computer vision},
  pages={7873--7883},
  year={2023}
}

@inproceedings{nakashima2022hyperbolic,
  title={Hyperbolic timbre embedding for musical instrument sound synthesis based on variational autoencoders},
  author={Nakashima, Futa and Nakamura, Tomohiko and Takamune, Norihiro and Fukayama, Satoru and Saruwatari, Hiroshi},
  booktitle={2022 Asia-Pacific Signal and Information Processing Association Annual Summit and Conference (APSIPA ASC)},
  pages={735--742},
  year={2022},
  organization={IEEE}
}

@article{li2025hlformer,
  title={Hlformer: Enhancing partially relevant video retrieval with hyperbolic learning},
  author={Li, Jun and Wang, Jinpeng and Tan, Chaolei and Lian, Niu and Chen, Long and Wang, Yaowei and Zhang, Min and Xia, Shu-Tao and Chen, Bin},
  journal={arXiv preprint arXiv:2507.17402},
  year={2025}
}

@inproceedings{li2025enhancing,
  title={Enhancing partially relevant video retrieval with hyperbolic learning},
  author={Li, Jun and Wang, Jinpeng and Tan, Chaolei and Lian, Niu and Chen, Long and Wang, Yaowei and Zhang, Min and Xia, Shu-Tao and Chen, Bin},
  booktitle={Proceedings of the IEEE/CVF International Conference on Computer Vision},
  pages={23074--23084},
  year={2025}
}

@inproceedings{chen2022fully,
  title={Fully hyperbolic neural networks},
  author={Chen, Weize and Han, Xu and Lin, Yankai and Zhao, Hexu and Liu, Zhiyuan and Li, Peng and Sun, Maosong and Zhou, Jie},
  booktitle={Proceedings of the 60th Annual Meeting of the Association for Computational Linguistics (Volume 1: Long Papers)},
  pages={5672--5686},
  year={2022}
}

@article{liu2019hyperbolic,
  title={Hyperbolic graph neural networks},
  author={Liu, Qi and Nickel, Maximilian and Kiela, Douwe},
  journal={Advances in neural information processing systems},
  volume={32},
  year={2019}
}

@misc{chen2020closer,
      title={A Closer Look at Few-shot Classification}, 
      author={Wei-Yu Chen and Yen-Cheng Liu and Zsolt Kira and Yu-Chiang Frank Wang and Jia-Bin Huang},
      year={2020},
      eprint={1904.04232},
      archivePrefix={arXiv},
      primaryClass={cs.CV},
      url={https://arxiv.org/abs/1904.04232}, 
}

@article{hou2019cross,
  title={Cross attention network for few-shot classification},
  author={Hou, Ruibing and Chang, Hong and Ma, Bingpeng and Shan, Shiguang and Chen, Xilin},
  journal={Advances in neural information processing systems},
  volume={32},
  year={2019}
}

@inproceedings{gu2022ppt,
  title={Ppt: Pre-trained prompt tuning for few-shot learning},
  author={Gu, Yuxian and Han, Xu and Liu, Zhiyuan and Huang, Minlie},
  booktitle={Proceedings of the 60th annual meeting of the association for computational linguistics (volume 1: long papers)},
  pages={8410--8423},
  year={2022}
}

@article{hu2023scaled,
  title={Scaled Prompt-Tuning for Few-Shot Natural Language Generation},
  author={Hu, Ting and Meinel, Christoph and Yang, Haojin},
  journal={arXiv preprint arXiv:2309.06759},
  year={2023}
}

@article{de2025take,
  title={Take a Peek: Efficient Encoder Adaptation for Few-Shot Semantic Segmentation via LoRA},
  author={De Marinis, Pasquale and Vessio, Gennaro and Castellano, Giovanna},
  journal={arXiv preprint arXiv:2512.10521},
  year={2025}
}

@article{alayrac2022flamingo,
  title={Flamingo: a visual language model for few-shot learning},
  author={Alayrac, Jean-Baptiste and Donahue, Jeff and Luc, Pauline and Miech, Antoine and Barr, Iain and Hasson, Yana and Lenc, Karel and Mensch, Arthur and Millican, Katherine and Reynolds, Malcolm and others},
  journal={Advances in neural information processing systems},
  volume={35},
  pages={23716--23736},
  year={2022}
}

@article{zhang2023multimodal,
  title={Multimodal chain-of-thought reasoning in language models},
  author={Zhang, Zhuosheng and Zhang, Aston and Li, Mu and Zhao, Hai and Karypis, George and Smola, Alex},
  journal={arXiv preprint arXiv:2302.00923},
  year={2023}
}

@article{hu2022context,
  title={In-context learning for few-shot dialogue state tracking},
  author={Hu, Yushi and Lee, Chia-Hsuan and Xie, Tianbao and Yu, Tao and Smith, Noah A and Ostendorf, Mari},
  journal={arXiv preprint arXiv:2203.08568},
  year={2022}
}

@inproceedings{cai2023context,
  title={In-context learning for few-shot multimodal named entity recognition},
  author={Cai, Chenran and Wang, Qianlong and Liang, Bin and Qin, Bing and Yang, Min and Wong, Kam-Fai and Xu, Ruifeng},
  booktitle={Findings of the Association for Computational Linguistics: EMNLP 2023},
  pages={2969--2979},
  year={2023}
}

@inproceedings{huang2024select,
  title={Select and order: Optimizing few-shot image classification with in-context learning},
  author={Huang, Hujiang and Xie, Yu and Gao, Jun and Fan, Chuanliu and Cao, Ziqiang},
  booktitle={International Conference on Multimedia Modeling},
  pages={425--438},
  year={2024},
  organization={Springer}
}

@article{nickel2017poincare,
  title={Poincar{\'e} embeddings for learning hierarchical representations},
  author={Nickel, Maximillian and Kiela, Douwe},
  journal={Advances in neural information processing systems},
  volume={30},
  year={2017}
}

@inproceedings{nickel2018learning,
  title={Learning continuous hierarchies in the lorentz model of hyperbolic geometry},
  author={Nickel, Maximillian and Kiela, Douwe},
  booktitle={International conference on machine learning},
  pages={3779--3788},
  year={2018},
  organization={PMLR}
}

@article{ren2024flowar,
  title={Flowar: Scale-wise autoregressive image generation meets flow matching},
  author={Ren, Sucheng and Yu, Qihang and He, Ju and Shen, Xiaohui and Yuille, Alan and Chen, Liang-Chieh},
  journal={arXiv preprint arXiv:2412.15205},
  year={2024}
}

@article{luo2025curveflow,
  title={Curveflow: Curvature-guided flow matching for image generation},
  author={Luo, Yan and Du, Drake and Huang, Hao and Fang, Yi and Wang, Mengyu},
  journal={arXiv preprint arXiv:2508.15093},
  year={2025}
}

@inproceedings{li2022grounded,
  title={Grounded language-image pre-training},
  author={Li, Liunian Harold and Zhang, Pengchuan and Zhang, Haotian and Yang, Jianwei and Li, Chunyuan and Zhong, Yiwu and Wang, Lijuan and Yuan, Lu and Zhang, Lei and Hwang, Jenq-Neng and others},
  booktitle=CVPR,
  pages={10965--10975},
  year={2022}
}

@inproceedings{zhong2022regionclip,
  title={Regionclip: Region-based language-image pretraining},
  author={Zhong, Yiwu and Yang, Jianwei and Zhang, Pengchuan and Li, Chunyuan and Codella, Noel and Li, Liunian Harold and Zhou, Luowei and Dai, Xiyang and Yuan, Lu and Li, Yin and others},
  booktitle=CVPR,
  pages={16793--16803},
  year={2022}
}

@inproceedings{pal2025compositional,
  title={Compositional entailment learning for hyperbolic vision-language models},
  author={Pal, Avik and van Spengler, Max and di Melendugno, Guido Maria D'Amely and Flaborea, Alessandro and Galasso, Fabio and Mettes, Pascal},
  booktitle=ICLR,
  year={2025}
}

@inproceedings{chenflow2024,
  title={Flow Matching on General Geometries},
  author={Chen, Ricky TQ and Lipman, Yaron},
  booktitle=ICLR,
year={2022}
}

@article{yang2024consistency,
  title={Consistency flow matching: Defining straight flows with velocity consistency},
  author={Yang, Ling and Zhang, Zixiang and Zhang, Zhilong and Liu, Xingchao and Xu, Minkai and Zhang, Wentao and Meng, Chenlin and Ermon, Stefano and Cui, Bin},
  journal={arXiv preprint arXiv:2407.02398},
  year={2024}
}

@inproceedings{li2024autoregressive,
  title={Autoregressive image generation without vector quantization},
  author={Li, Tianhong and Tian, Yonglong and Li, He and Deng, Mingyang and He, Kaiming},
  booktitle=nips,
  year={2024}
}

@article{li2025text,
  title={H2em: Learning Hierarchical Hyperbolic Embeddings for Compositional Zero-Shot Learning},
  author={Li, Lin and Li, Jiahui and Lei, Jiaming and Xiao, Jun and Shao, Feifei and Chen, Long},
  journal={arXiv preprint arXiv:2512.20029},
  year={2025}
}
\bibliographystyle{icml2026}

\newpage
\appendix

\end{document}